\documentclass[10pt,twocolumn,letterpaper]{article}

\usepackage[dvipsnames]{xcolor}
\usepackage{subcaption}
\usepackage{cvpr}
\usepackage{times}
\usepackage{graphicx}
\usepackage{epstopdf}
\usepackage{amsmath}
\usepackage{amssymb}

\usepackage{amsthm}
\usepackage{enumitem}
\usepackage{empheq}
\usepackage{mathtools}
\usepackage{textcomp}
\usepackage{booktabs}
\usepackage{fixltx2e}
\usepackage{multirow}
\usepackage{makecell}
\usepackage{svg}

\graphicspath{{resources/}}
\newcommand{\diag}{\mathop{\operatorname{diag}}}

\newcommand{\cov}{\mathop{\operatorname{Cov}}\nolimits}
\newcommand{\var}{\mathop{\operatorname{Var}}\nolimits}

\let\originalleft\left
\let\originalright\right
\renewcommand{\left}{\mathopen{}\mathclose\bgroup\originalleft}
\renewcommand{\right}{\aftergroup\egroup\originalright}
\DeclareMathOperator*{\argmin}{arg\,min}

\definecolor{box_red}{RGB}{240, 60, 10}
\definecolor{box_green}{RGB}{10, 180, 10}
\definecolor{box_cyan}{RGB}{10, 160, 160}
\definecolor{box_blue}{RGB}{10, 10, 210}

\makeatletter
\renewcommand{\paragraph}{
  \@startsection{paragraph}{4}
  {\z@}{0.7ex \@plus .2ex \@minus .1ex}{-1em}
  {\normalfont\normalsize\bfseries}
}
\renewcommand{\subparagraph}{
  \@startsection{subparagraph}{5}
  {\parindent }{0ex \@plus .2ex \@minus 0ex}{-1em}
  {\normalfont \normalsize \bfseries }
}
\makeatother

\usepackage[pagebackref=true,breaklinks=true,letterpaper=true,colorlinks,bookmarks=false]{hyperref}
\usepackage{cleveref}

\cvprfinalcopy 

\def\cvprPaperID{7023}

\begin{document}

\title{
MonoRUn: Monocular 3D Object Detection by Reconstruction and \protect\\ Uncertainty Propagation
}

\author{
Hansheng Chen, 
Yuyao Huang,
Wei Tian\thanks{Corresponding author: Wei Tian.}, 
Zhong Gao,
Lu Xiong\\
Institute of Intelligent Vehicles, School of Automotive Studies, Tongji University\\
{\tt\small
hanshengchen97@gmail.com \{huangyuyao, tian\_wei, 1931604, xiong\_lu\}@tongji.edu.cn
}
}

\maketitle

\begin{abstract}
Object localization in 3D space is a challenging aspect in monocular 3D object detection. Recent advances in 6DoF pose estimation have shown that predicting dense 2D-3D correspondence maps between image and object 3D model and then estimating object pose via Perspective-n-Point (PnP) algorithm can achieve remarkable localization accuracy. Yet these methods rely on training with ground truth of object geometry, which is difficult to acquire in real outdoor scenes. To address this issue, we propose MonoRUn, a novel detection framework that learns dense correspondences and geometry in a self-supervised manner, with simple 3D bounding box annotations. To regress the pixel-related 3D object coordinates, we employ a regional reconstruction network with uncertainty awareness. For self-supervised training, the predicted 3D coordinates are projected back to the image plane. A Robust KL loss is proposed to minimize the uncertainty-weighted reprojection error. During testing phase, we exploit the network uncertainty by propagating it through all downstream modules. More specifically, the uncertainty-driven PnP algorithm is leveraged to estimate object pose and its covariance. Extensive experiments demonstrate that our proposed approach outperforms current state-of-the-art methods on KITTI benchmark.\footnote{Code: \url{https://github.com/tjiiv-cprg/MonoRUn}.}
\end{abstract}

\begin{figure}[t]
\begin{center}
   \includegraphics[width=0.92\linewidth]{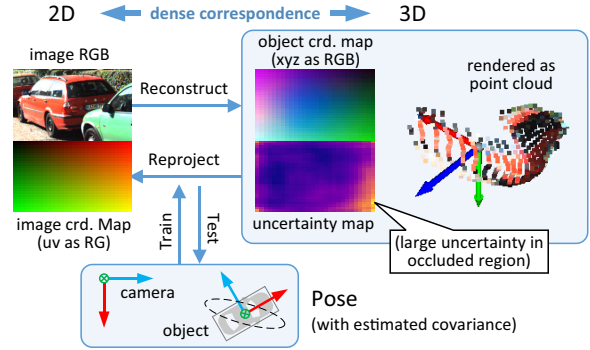}
\end{center}
   \vspace{-2mm}
   \caption{3D reconstruction is conducted by regressing the pixel-related object coordinate map, which can be visualized as local point cloud in object space. For self-supervision, the object coordinates are reprojected to recover the image coordinate map. To focus on foreground pixels, we estimate the aleatoric uncertainty of network prediction. The coordinate uncertainty can be further propagated to estimate the pose covariance.}
\label{fig:abstract}
\end{figure}

\section{Introduction}
Monocular 3D object detection is an active research area in computer vision. Although deep-learning-based 2D object detection has achieved remarkable progress~\cite{cascade, faster}, the 3D counterpart still poses a much greater challenge on accurate object localization, since a single image cannot provide explicit depth information. To address this issue, a large number of works leverage geometrical priors and solve the object pose (position and orientation in camera frame) via 2D-3D constraints. These constraints either require extra keypoint annotations~\cite{deep-manta, mono3d++}, or exploit centers, corners and edges of ground truth bounding boxes~\cite{RTM3D, 3dbbox}. 
Yet the accuracy largely depends on the number and quality of available constraints, and the performance degrades in occlusion and truncation cases with fewer visible keypoints. 
A more robust approach is using dense 2D-3D correspondences, in which every single foreground pixel is mapped to a 3D point in the object space. This 
has proven successful in monocular 6DoF pose estimation tasks~\cite{linemod}.

Current state-of-the-art dense correspondence methods~\cite{CDPN, pix2pose, DPOD} rely on both ground truth pose and 3D object model, so that target 3D coordinate map and object mask can be rendered for training supervision. This requirement restricts the training data to synthetic or simple laboratory scenes, where exact or pre-reconstructed 3D models are readily available. However, 3D object detection in real scenes (\eg, driving scenes) mostly deals with category-level objects, where acquiring accurate 3D models for all instances is impractical. An intuitive idea could be using LiDAR points to generate sparse coordinate maps for supervision. However, the persisting challenge is the deficiency of LiDAR points on specific objects or parts, \eg, on distant objects or reflective materials. 

A workaround to the lack of ground truth is leveraging self-supervision. Typically, Wang \etal~\cite{Self6D} adopted a self-supervised network to directly learn object pose with the given ground truth 3D geometry. Our work, on the contrary, adopts the opposite idea: learning the 3D geometry from ground truth pose in a self-supervised manner during training, and then solving the object pose via 2D-3D correspondences during testing. 

In this paper, we propose the \textit{MonoRUn}, a novel \textbf{mono}cular 3D object detection method using self-supervised \textbf{r}econstruction with \textbf{un}certainty propagation. MonoRUn can extend off-the-shelf 2D detectors by appending a 3D branch to the region of interest (RoI) within each predicted 2D box. The 3D branch regresses dense 3D object coordinates in the RoI, which effectively builds up the geometry and 2D-3D correspondences. 

To overcome the need for supervised foreground segmentation, we estimate the uncertainty of the predicted coordinates and adopt an uncertainty-driven PnP algorithm, which focuses on the low-uncertainty foreground. Furthermore, by forward propagating the uncertainty through PnP module, we can estimate the pose covariance matrix, which is used for scoring the detection confidence.

Self-supervision is conducted by projecting the predicted 3D coordinates back to the image via ground truth object pose and camera intrinsic parameters. To minimize the reprojection error with uncertainty awareness, we propose the \textit{Robust KL loss} that minimizes the KL divergence between the predicted Gaussian distribution and the ground truth Dirac distribution. This novel loss function is the key to the state-of-the-art performance for the MonoRUn network.

To summarize, our main contributions are as follows: 
\begin{itemize}[noitemsep,topsep=0.7ex]
    \item We propose a novel monocular 3D object detection network with uncertainty awareness,
    which can be trained without extra annotations (\eg, keypoints, 3D models, masks). To the best of our knowledge, this is the first dense correspondence method employed for 3D detection in real driving scenes. 
    \item We propose the Robust KL loss for general deep regression with uncertainty awareness, and demonstrate its superiority over the plain KL loss in previous work.
    \item Extensive evaluation on KITTI~\cite{kitti} benchmark shows significant improvement of our approach compared to current state-of-the-art methods.
\end{itemize}

\section{Related Work}
\paragraph{Monocular 3D Object Detection} 
The majority of previous methods can be roughly divided into the following two categories, based on how depth information is derived.

\textbf{1) With off-the-shelf monocular depth estimators}. Representatively, Pseudo-LiDAR~\cite{Pseudo-LiDAR} converts the estimated depth map to 3D point cloud and takes advantage of existing LiDAR-based 3D object detection pipeline.
D\textsuperscript{4}LCN~\cite{D4LCN} uses depth map as guidance to generate dynamic depth-wise convolutional filters, which can extract 3D information from RGB image more effectively. These methods largely benefit from the pre-trained depth estimator, \eg, DORN~\cite{DORN}, which may have generalization issues.

\textbf{2) With 2D-3D geometrical constraints}. Deep-MANTA~\cite{deep-manta} annotates the training data with 36-keypoint 3D vehicle templates. 
A network is trained to find the the best-matched template and regress 2D keypoint coordinates, and vehicle pose is computed via EPnP~\cite{EPnP} algorithm. 
RTM3D~\cite{RTM3D} detects virtual keypoints (corners and center of the 3D bounding box) using a CenterNet-like~\cite{centernet} network. 
Another commonly used constraint is 2D box and 3D box consistency, 
first used by Mousavian \etal~\cite{3dbbox}. The above methods suffer from constraint deficiency by truncation or occlusion.

\paragraph{Dense Correspondence and 3D Reconstruction} Most existing work uses ground truth geometry to train deep correspondence mapping networks. Nevertheless, some have explored end-to-end training via differentiable PnP algorithm without ground truth geometry.

\textbf{1) With geometry supervision}. Pix2Pose~\cite{pix2pose} directly regresses normalized object coordinates (NOC) of each object pixel.
DPOD~\cite{DPOD} predicts two-channel UV correspondences that map the object surface to 3D coordinates. Regarding category-level objects, Wang \etal~\cite{NOCS} demonstrated that the scale-invariant NOC can handle unseen instances in a given category. 
These methods are only tested on synthetic or simple indoor data.

\textbf{2) Without geometry supervision}. Brachmann and Rother~\cite{dsac++} proposed an approximate PnP back-propagation method to train an end-to-end network for Structure from Motion (SfM) problem.
A further development is the BPnP~\cite{BPnP}, which is an exact PnP back-propagation approach. However, both methods train the networks in conjunction with reprojection loss as regularization term, which is essentially self-supervision. The potential of self-supervision alone is not investigated in these researches.

\begin{figure*}[t]
   \begin{center}
   \includegraphics[width=0.88\textwidth]{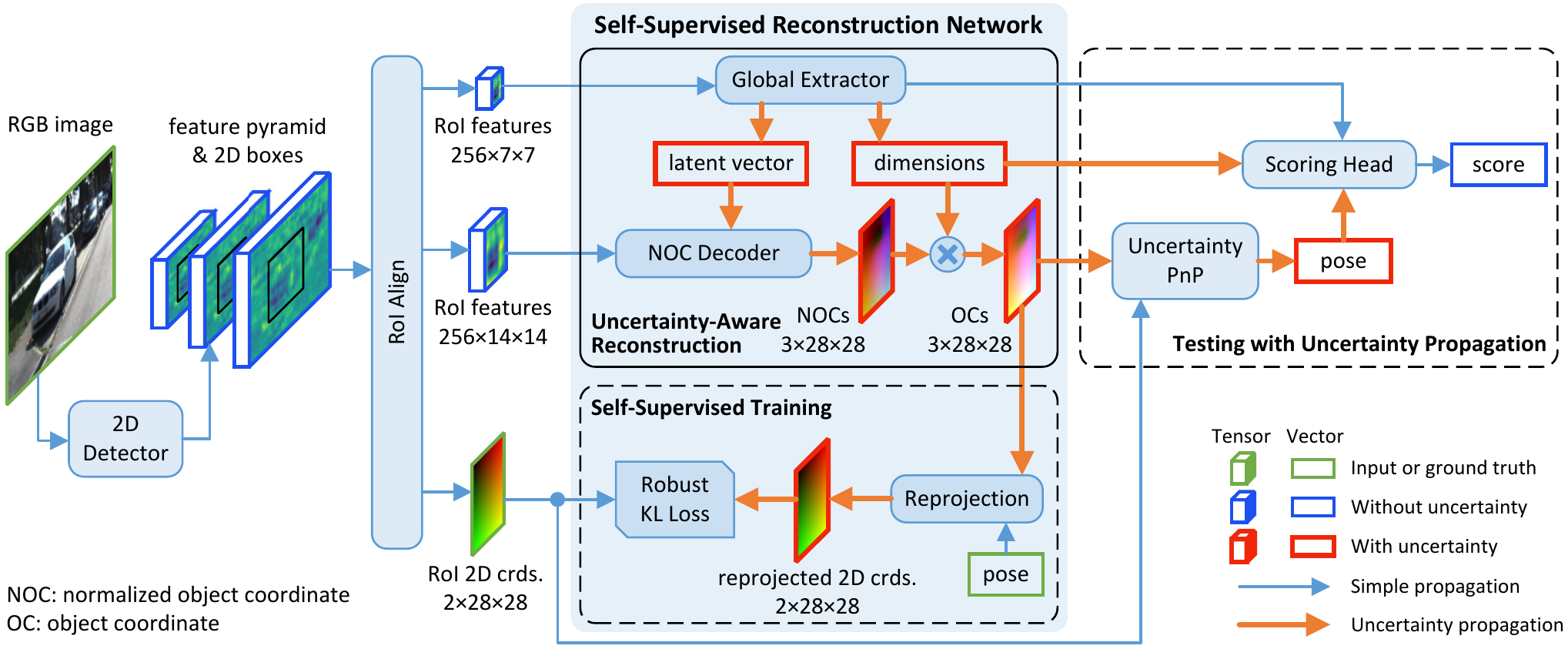}
   \end{center}
   \vspace{-2mm}
   \caption{\textbf{Training and testing pipeline of MonoRUn.} The uncertainty-aware variables in \textcolor{red}{red} are modeled by Monte Carlo approach or by probabilistic models (\eg, Gaussian models).} 
\label{fig:overview}
\end{figure*}

\paragraph{Uncertainty Estimation} 
The uncertainty in deep learning consists of \textit{aleatoric} and \textit{epistemic} uncertainty~\cite{kendall2017uncertainties}. The former captures noise inherent in observations, while the latter represents the uncertainty of model parameters.
Kendall and Gal~\cite{kendall2017uncertainties} introduced heteroscedastic regression for deep networks to directly output data-dependent aleatoric uncertainty, which can be learned with a \textit{KL loss}~\cite{klloss, kendall2017uncertainties} function. Yet the plain KL loss is sensitive to outliers and is not well-balanced against other loss functions, leaving room for improvement.

\section{Proposed Approach}
\subsection{Problem Formulation and Approach Overview}
Given an RGB image, the aim of 3D object detection is to localize and classify all objects of interest, yielding a 3D bounding box with class label for each object. The 3D box can be parameterized with dimensions $\mathbf{d} = \begin{bmatrix} l & h & w \end{bmatrix}^\text{T}$ and pose $\mathbf{p} = \begin{bmatrix} \beta & t_\text{x} & t_\text{y} & t_\text{z} \end{bmatrix}^\text{T}$, where $\beta$ is the object yaw angle and $t_\text{x}, t_\text{y}, t_\text{z}$ is the bottom center of the box in camera coordinate system.

Based on off-the-shelf 2D object detectors, we aim to predict a 3D object coordinate map using the RoI features within each 2D box. For self-supervision, given the ground truth pose $\mathbf{p}_\text{gt}$ and camera model, we can project the 3D coordinates back to the image, obtaining the reprojected 2D coordinates $(u_\text{rp}, v_\text{rp})$. The objective is to recover the original 2D coordinates $(u, v)$. However, if we simply minimize the reprojection error without foreground segmentation, the network will be disrupted by large errors (hence large uncertainty) on irrelevant background points. Therefore, we design an \textit{uncertainty-aware reconstruction} module that estimates the aleatoric uncertainty of $(u_\text{rp}, v_\text{rp})$, and the network is optimized by minimizing the uncertainty-weighted reprojection error using the proposed Robust KL loss.

During inference, we adopt the \textit{uncertainty-driven PnP} module, through which the network uncertainty is propagated to object pose, represented with a multivariate normal distribution. This distribution is used by the scoring head to compute the detection confidence.

\subsection{Self-Supervised Reconstruction Network}

To deal with category-level objects of various sizes, we employ two network branches to predict the 3D dimensions and the dimension-invariant normalized object coordinates (NOC)~\cite{NOCS} respectively. Then, the object coordinate vector $\mathbf{x}^\text{OC}$ is the element-wise product (denoted by $\circ$) of NOC vector $\mathbf{x}^\text{NOC}$ and dimension vector $\mathbf{d}$:
\begin{equation}
    \mathbf{x}^\text{OC} = \mathbf{x}^\text{NOC} \circ \mathbf{d}.
\end{equation}
The first branch is called the \textit{global extractor}, which extracts the global understanding of an object and predicts the 3D dimensions. The second branch is called the \textit{NOC decoder}, which predicts the dense NOC map using convolutional layers. Since convolutional layers have limited capability of understanding the global context, we let the global extractor predict an additional global latent vector to enhance the NOC decoder. This latent vector potentially encodes the occlusion, truncation and shape of the object, and is found to be beneficial for aleatoric uncertainty estimation (as shown in Section~\ref{ablation}). Details about the network are presented as follows. 

\begin{figure}[t]
\begin{center}
    \includegraphics[width=0.95\linewidth]{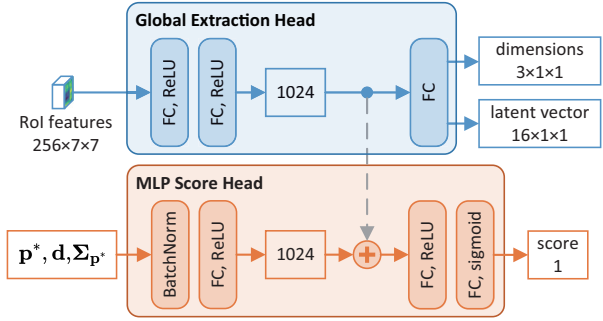}
\end{center}
   \vspace{-2mm}
   \caption{\textbf{The global extraction head} predicts the dimensions and latent vector, while \textbf{the MLP score head} exploits both the detection results and the global feature to predict the detection score.}
\label{fig:shapescore}
\end{figure}

\paragraph{Global Extractor} As shown in Fig.~\ref{fig:overview}, a 7\texttimes7 RoI feature map is extracted from a higher level of the feature pyramid. The features are then flattened and fed into the global extraction head, which extracts a 16-channel global latent vector and predicts the 3D dimensions. The dimensions can be directly supervised by the annotated 3D bounding box sizes. For this network branch, we adopt two 1024-channel fully connected layers, as shown in Fig.~\ref{fig:shapescore}.

\begin{figure}[t]
\begin{center}
    \includegraphics[width=0.97\linewidth]{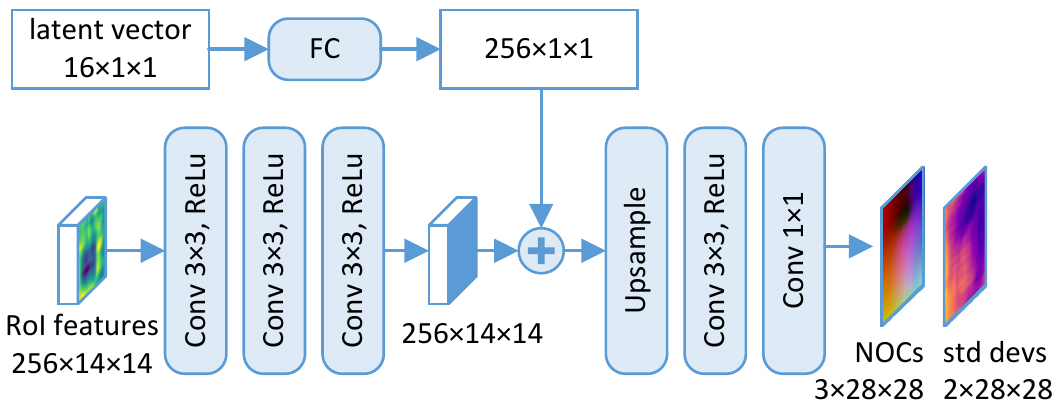}
\end{center}
   \vspace{-2mm}
   \caption{\textbf{The NOC decoder network} aggregates the local features and global embeddings by channel-wise addition, and outputs the dense NOCs as well as standard deviations of reprojected coordinates. For upsampling, we adopt the CARAFE layer~\cite{carafe}, which is more efficient than the deconvolutional layer often used in instance segmentation. }
\label{fig:nocdecoder}
\end{figure}

\paragraph{NOC Decoder} The NOC decoder network is designed to aggregate the global latent vector and local convolutional features for NOC prediction. 
This is realized by incorporating the \textit{Excitation} operation from the Squeeze-Excitation Network~\cite{SENet}. As shown in Fig.~\ref{fig:nocdecoder}, the channel size of the latent vector is first expanded to 256. Then, a channel-wise addition is conducted before the upsampling layer. Apart from predicting the three-channel NOC map, the NOC decoder network is also responsible for estimating the two-channel aleatoric uncertainty map, as explained in the next paragraph.

\paragraph{Self-Supervision with Aleatoric Uncertainty} Given the ground truth of object pose, the predicted object coordinates can be projected back to the image plane. The reprojection error of the pixel $(u, v)$ is formulated as:
\begin{equation}
\mathbf{r}_{(u, v)} = K(\mathbf{R} \mathbf{x}^\text{OC}_{(u, v)} + \mathbf{t}) - \begin{bmatrix} u \\ v \end{bmatrix}
\label{eqn:repoj}
\end{equation}
with the camera projection function $K(\cdot)$, rotation matrix $\mathbf{R}$, and translation vector $\mathbf{t}=\begin{bmatrix} t_\text{x} & t_\text{y} & t_\text{z} \end{bmatrix}^\text{T}$. 
To focus on minimizing foreground error without instance segmentation, we introduce the aleatoric uncertainty in our work. More specifically, we use univariate Gaussian distribution to model the reprojected 2D coordinates, and let the network predict the means and standard deviations, which are optimized using the Robust KL loss. 
Formally we could follow the uncertainty propagation path in Fig.~\ref{fig:overview} and predict the uncertainty of intermediate variables at first. Practically, leveraging the flexibility of deep networks, we take a shortcut by letting the NOC decoder directly regress the standard deviations of reprojected 2D coordinates, as shown in Fig.~\ref{fig:nocdecoder}. 

\paragraph{Additional Epistemic Uncertainty} Estimating the epistemic uncertainty is essential for safety-critical applications such as autonomous driving.
Following \cite{kendall2017uncertainties}, we compute the mean and variance of $\mathbf{x}^\text{OC}$ using Monte Carlo dropout during inference. We insert a channel dropout layer~\cite{channeldropout} after RoI Align~\cite{maskrcnn} and a 1D dropout layer after each FC layer. Since PnP algorithms handle 2D projection variances more efficiently, we first transform the 3D variances of object coordinates into 2D variances of reprojected coordinates, which is then combined with the aleatoric uncertainty. Details are shown in the supplementary materials.

\subsection{Robust KL Loss} \label{rkl}
By definition, KL loss is derived from the KL divergence between the predicted distribution and target distribution. Assuming Gaussian priors, the KL divergence is as follows:
\begin{multline}
D_\text{KL}(\mathcal{N}_\text{tgt} \| \mathcal{N}_\text{pred}) \\ = \frac{1}{2} \left( \frac{\sigma_\text{tgt}^2}{\sigma_\text{pred}^2\negthickspace} + \frac{(\mu_\text{pred} - \mu_\text{tgt})^2}{\sigma_\text{pred}^2} - 1 + \log\frac{\sigma_\text{pred}^2}{\sigma_\text{tgt}^2} \right).
\end{multline}
For fixed target distribution, $\log{\sigma_\text{tgt}^2}$ is constant and can be omitted in the minimization. Assuming narrow target (Dirac-like), $(\mu_\text{pred} - \mu_\text{tgt})^2$ dominates much more than $\sigma_\text{tgt}^2$. Let $y = \mu_\text{tgt}$, the minimization objective is simplified as:
\begin{equation}
L_\text{KL} = \frac{1}{2 \sigma_\text{pred}^2\negthickspace} (\mu_\text{pred} - y)^2 + \frac{1}{2} \log{\sigma_\text{pred}^2}.
\label{eqn:klloss}
\end{equation}
We call Eq.~(\ref{eqn:klloss}) the Gaussian KL loss. Hereafter we omit the subscript $(\cdot)_\text{pred}$ for brevity. To capture heteroscedastic aleatoric uncertainty in regression, Kendall and Gal~\cite{kendall2017uncertainties} proposed to directly predict the data-dependent mean $\mu$ and log variance $\log \sigma^2$ using deep networks, which is optimized by Eq.~(\ref{eqn:klloss}). Apparently, the first term in Eq.~(\ref{eqn:klloss}) is a weighted L2 loss, where errors with higher uncertainty are less punished. 

Despite its probabilistic origin, the Gaussian KL loss has two defects when applied to a deep regression model: 
\begin{itemize}[noitemsep,topsep=0.7ex]
    \item As a generalization form of L2 loss, the Gaussian KL loss is not robust to outliers;
    \item The gradient \wrt $\mu$ tends to increase as the denominator $2\sigma^2$ decays during training, whereas regular L2 or L1 losses have decreasing or constant gradient, leading to loss imbalance in multi-task learning.
\end{itemize}

Regarding the first problem, an alternative KL loss derived from Laplacian distribution is used in work~\cite{monopair, kendall2017uncertainties}:
\begin{equation}
    L_\text{LapKL} = \frac{\sqrt{2}}{\sigma} |\mu - y| + \log{\sigma}.
    \label{eqn:lapkl}
\end{equation}
Same as the L1 loss, this function is not differentiable at $\mu = y$. To overcome this issue, we design a mixed KL loss, written as a function of weighted error $e=(\mu - y) / \sigma$ and standard deviation $\sigma$:
\begin{equation}
 L_\text{mKL} = 
 \begin{dcases}
 \frac{1}{2} e^2 + \log{\sigma}, & |e| \leq \sqrt{2},\\
 \sqrt{2}|e| - 1 + \log{\sigma}, & |e| > \sqrt{2}.
 \end{dcases}
\label{eqn:mklloss}
\end{equation}
It can be verified that this function is differentiable \wrt both $\mu$ and $\sigma$ (on condition that $\sigma > 0$).
The mixed KL loss can be regarded as an extension of Huber loss (smooth L1), which is robust to outliers and easier to optimize.

The second problem is caused by the increasing weight $1 / \sigma$ as the denominator $\sigma$ decays during training. This can be mitigated by normalizing the weight such that its batch-wise mean equals unity. Inspired by Batch Normalization~\cite{batchnorm}, we perform online estimate of the mean weight $\hat{w} = \text{E}[1 / \sigma]$ using exponential moving average:
\begin{equation}
    \hat{w} \gets \alpha \hat{w} + (1 - \alpha) \frac{1}{N} \sum_{i=1}^{N}{\frac{1}{\sigma_i}},
\end{equation}
where $\alpha$ is the momentum, $N$ is the number of samples in a batch. The finalized Robust KL loss is simply the weight-normalized mixed KL loss:
\begin{equation}
 L_\text{RKL} = \frac{1}{\hat{w}} L_\text{mKL} .
\end{equation}

In practice, directly optimizing $\sigma$ can lead to gradient explosion. Therefore, we let the network predict the logarithmic standard deviation $\log{\sigma}$ as an alternative.

\subsection{Uncertainty-Driven PnP}
\paragraph{Maximum Likelihood Estimation}
To solve PnP with uncertainty is to perform the maximum likelihood estimation (MLE) of pose $\mathbf{p}$, in which the negative log likelihood (NLL) function is the sum of squared reprojection error $\mathbf{r}_{(u, v)}$ measured in Mahalanobis distance:
\begin{equation}
\mathbf{p}^* = 
\argmin_\mathbf{p} \frac{1}{2} \negthickspace \negthickspace \negthickspace \smashoperator[r]{\sum_{(u, v) \in RoI}}
\mathbf{r}_{(u, v)}^\text{T} \mathbf{\Sigma}_{(u, v)}^{-1} \mathbf{r}_{(u, v)}
\label{eqn:mle}
\end{equation}
with $\mathbf{\Sigma}_{(u, v)} = \diag[\sigma_{u_\text{rp}}^2, \sigma_{v_\text{rp}}^2] \negmedspace \bigm|_{(u, v)}$, where $\sigma_{u_\text{rp}}, \sigma_{v_\text{rp}}$ are the predicted standard deviations of the reprojected coordinates. This minimization can be solved efficiently using the Levenberg-Marquardt algorithm.

\paragraph{Covariance Estimation}
For probabilistic object localization, we also need to estimate the covariance of $\mathbf{p}^*$, which can be approximated by the inverse Hessian of the NLL at $\mathbf{p}^*$~\cite{likelihood}:
\begin{equation}
    \cov[\mathbf{p}^*] 
    \approx \mathbf{H}(\mathbf{p}^*)^{-1}.
\label{eqn:cov}
\end{equation}
To avoid computing the second derivative during inference, we approximate the exact Hessian using the Gauss-Newton matrix $\mathbf{J}(\mathbf{p}^*)^\text{T} \mathbf{J}(\mathbf{p}^*)$, with $\mathbf{J}(\mathbf{p}^*) = \partial \mathbf{r}_\text{all} / \partial \mathbf{p}^\text{T} \negthickspace \bigm|_{\mathbf{p}^*}$, where $\mathbf{r}_\text{all} = \begin{bmatrix} r_1 / \sigma_1 & r_2 / \sigma_2 & \dots & r_{2n} / \sigma_{2n} \end{bmatrix}^\text{T}$ (flattened vector of all weighted reprojection errors). 

\paragraph{Online Covariance Calibration}
In practice, using Eq.~(\ref{eqn:cov}) can result in a much lower covariance than the actual one. This is mainly because Eq.~(\ref{eqn:mle}) assumes that the reprojection error of each point is independent, whereas the network outputs are apparently correlated. Therefore, we further conduct online covariance calibration using a 4\texttimes1 learnable calibration vector $\mathbf{k}$:
\begin{equation}
    \mathbf{\Sigma}_{\mathbf{p}^*} = \exp(\diag \mathbf{k}) \left(\mathbf{J}(\mathbf{p}^*)^\text{T} \mathbf{J}(\mathbf{p}^*)\right)^{-1} \negthickspace \exp(\diag \mathbf{k}).
\end{equation}
The calibration vector can be learned by applying the multivariate Gaussian KL loss:
\begin{equation}
    L_\text{calib} = \frac{1}{2} (\mathbf{p}^*\negthickspace - \mathbf{p}_\text{gt})^\text{T} \mathbf{\Sigma}_{\mathbf{p}^*}^{-1} (\mathbf{p}^*\negthickspace - \mathbf{p}_\text{gt}) + \frac{1}{2} \log \det \mathbf{\Sigma}_{\mathbf{p}^*},
\label{eqn:calibloss}
\end{equation}
where $\mathbf{p}^*$ is detached and only the calibration vector is optimized. Despite the defects of Gaussian KL loss stated in Section~\ref{rkl}, it is sufficient for the simple calibration task. 

\subsection{Scoring Head}
The confidence score of a detected object, denoted by $c$, 
can be decomposed into localization score $\Pr[F\negthinspace g]$ (the probability of detecting a foreground object) and classification score $\Pr[Cls]$ (the probability of predicting the correct class label). For 3D object detection, the score can be expressed as the product of 3D localization conditional probability and 2D score:
\begin{equation}
c_\text{3D} = \underbrace{\Pr[F\negthinspace g_\text{3D} | F\negthinspace g_\text{2D}]}_{\textstyle c_\text{3DLoc}} \underbrace{\Pr[F\negthinspace g_\text{2D}] \Pr[Cls]}_{\textstyle c_\text{2D}}.
\end{equation}
The 2D score $c_\text{2D}$ is given by the 2D detection module. Thus, we only need to predict the 3D localization score $c_\text{3DLoc}$. Since the 3D branch is trained only with positive 2D samples, the predicted $c_\text{3DLoc}$ is naturally conditional.

By sampling object poses from the estimated distribution $\mathcal{N}(\mathbf{p}^*, \cov[\mathbf{p}^*])$, the 3D localization score can be computed via Monte Carlo integration \wrt 3D IoU, as we show in the supplementary material. Due to slow 3D IoU calculation, Monte Carlo scoring has adverse effects on inference time. Therefore, we practically adopt the multi-layer perceptron (MLP) approach (Fig.~\ref{fig:shapescore}), which is faster, end-to-end trainable, and capable of fusing both pose uncertainty and network features to predict a more reliable score. To train the MLP scoring branch, we use the same binary cross-entropy loss as in \cite{pvrcnn, parta2}:
\begin{equation}
    L_\text{score} =
    -c_\text{tgt} \log c_\text{3DLoc} - (1 - c_\text{tgt}) \log(1 - c_\text{3DLoc}),
\end{equation}
where $c_\text{3DLoc}$ is the output of MLP, which is bounded to 0\texttildelow1 by logistic activation function, $c_\text{tgt}$ is a clamped linear function \wrt the 3D IoU between prediction and ground truth:
\begin{equation}
    c_\text{tgt} = \max(0, \min(1, 2 I\negthinspace oU_\text{3D} - 0.5)).
\end{equation}
The performance comparison between Monte Carlo and MLP is presented in the supplementary materials.

\subsection{Network Training}
The proposed MonoRUn network can be trained with three different setups, which are compared in the experiments.

\paragraph{Fully Self-Supervised Reconstruction (Without Extra Supervision)} In this mode, neither the LiDAR point supervision nor the end-to-end PnP is used. The 3D reconstruction process is trained in a fully self-supervised manner, except that 3D dimensions are directly supervised. The overall loss function is formulated as:
\begin{equation}
    L_\text{self} = L_\text{2D} + L_\text{proj} + L_\text{dim} + L_\text{score} + \lambda L_\text{calib},
\end{equation}
where $L_\text{2D}$ is the 2D detection loss, $L_\text{proj}$ is a Robust KL loss on self-supervised reprojection error, $L_\text{dim}$ is a smooth L1 loss on dimension error, and $\lambda$ is a hyperparameter for calibration loss, which is set to 0.01. 

\paragraph{LiDAR Supervision} Direct NOC loss can be imposed by converting foreground LiDAR points into sparse ground truth of NOC map. In this case, the aleatoric uncertainty is unnecessary. Thus, we adopt the weighted smooth L1 loss:
\begin{equation}
    L_\text{NOC} = \frac
    {1} {\sum\limits_i w_i} \sum_i w_i L_\text{SmoothL1}(x^\text{NOC}_{i, \text{pred}} - x^\text{NOC}_{i, \text{LiDAR}}),
\end{equation}
where $x^\text{NOC}_i$ denotes the $i$-th element of the NOC tensor, $w_i$ equals 1 where $x^\text{NOC}_{i, \text{LiDAR}}$ is available and 0 elsewhere. The overall loss becomes:
\begin{equation}
    L_\text{LiDAR} = L_\text{self} + L_\text{NOC}.
\end{equation}
Without specific statement, we train all the models using this setup.

\paragraph{End-to-End Training} Incorporating the PnP back-propagation approach in \cite{BPnP}, we apply smooth L1 loss on the Euclidean errors of estimated translation vector and yaw angle. Details of the PnP derivatives and loss functions are presented in the supplementary materials. Since end-to-end training is unstable at the beginning, we use a similar training protocol to~\cite{dsac++}, \ie, applying end-to-end training as refinement after self-supervised training. \textbf{This setup is only investigated in ablation studies for pure comparison.}

\begin{table*}[t]
    \begin{center}
    \scalebox{0.8}{%
    \setlength{\tabcolsep}{0.8em}
    \begin{tabular}{lcccccccccc}
        \toprule
        \multirow{2}[1]{*}{Method} & \multicolumn{3}{c}{Test AP$_{IoU\geq0.7}$} & \multicolumn{3}{c}{Val AP$_{IoU\geq0.5}$} & \multicolumn{3}{c}{Val AP$_{IoU\geq0.7}$} & \multirow{2}[2]{*}{\shortstack{Time \\ (sec)}} \\ \cmidrule(lr){2-4} \cmidrule(lr){5-7} \cmidrule(lr){8-10}
        {} & Easy & Mod. & Hard & Easy & Mod. & Hard & Easy & Mod. & Hard \\
        \midrule
        M3D-RPN~\cite{m3drpn} & 14.76 & 9.71 & 7.42 & 48.53 & 35.94 & 28.59 & 14.53 & 11.07 & 8.65 & 0.16 \\ 
        MonoPair~\cite{monopair} & 13.04 & 9.99 & 8.65 & \textcolor{RoyalBlue}{55.38} & \textcolor{orange}{42.39} & \textcolor{orange}{37.99} & \textcolor{RoyalBlue}{16.28} & \textcolor{orange}{12.30} & \textcolor{orange}{10.42} & 0.057 \\
        RTM3D~\cite{RTM3D} & 14.41 & 10.34 & 8.77 & - & - & - & - & - & - & 0.055 \\
        AM3D*~\cite{AM3D} & \textcolor{RoyalBlue}{16.50} & 10.74 & \textcolor{RoyalBlue}{9.52} & - & - & - & \textcolor{gray}{28.31} & \textcolor{gray}{15.76} & \textcolor{gray}{12.24} & 0.4 \\
        PatchNet*~\cite{patchnet} & 15.68 & \textcolor{RoyalBlue}{11.12} & \textcolor{orange}{10.17} & - & - & - & \textcolor{gray}{31.6} & \textcolor{gray}{16.8} & \textcolor{gray}{13.8} & 0.4 \\
        D\textsuperscript{4}LCN*~\cite{D4LCN} & \textcolor{orange}{16.65} & \textcolor{orange}{11.72} & 9.51 & - & - & - & \textcolor{gray}{22.32} & \textcolor{gray}{16.20} & \textcolor{gray}{12.30} & 0.2
        \\ \midrule
        Ours (w/o extra supv.) & 16.04 & 10.53 & 9.11 & \textcolor{orange}{55.88} & \textcolor{RoyalBlue}{40.03} & \textcolor{RoyalBlue}{35.59} & \textcolor{orange}{17.26} & \textcolor{RoyalBlue}{12.27} & \textcolor{RoyalBlue}{10.41} & 0.070 \\ 
        Ours (+ LiDAR supv.) & \textcolor{red}{19.65} & \textcolor{red}{12.30} & \textcolor{red}{10.58} & \textcolor{red}{59.71} & \textcolor{red}{43.39} & \textcolor{red}{38.44} & \textcolor{red}{20.02} & \textcolor{red}{14.65} & \textcolor{red}{12.61} & 0.070
        \\ \bottomrule
    \end{tabular}}
    \end{center}
    \vspace{-1.6mm}
    \caption{\textbf{3D detection performance of Car category} on KITTI official test set and validation set. The \textcolor{red}{1\textsuperscript{st}}, \textcolor{orange}{2\textsuperscript{nd}} and \textcolor{RoyalBlue}{3\textsuperscript{rd}} place are color coded. * indicates using the pre-trained depth estimator DORN~\cite{DORN}. Wang \etal~\cite{wang2020foresee} pointed out that the training data of DORN overlaps with KITTI-Object validation data, causing the 3D detectors to overfit. We use \textcolor{gray}{gray} to indicate the values affected by overfitting.}
    \label{tab:compare}
\end{table*}

\begin{table}[t]
    \vspace*{-2mm}
    \begin{center}
    \scalebox{0.8}{%
    \setlength{\tabcolsep}{0.55em}
    \begin{tabular}{lcccccc}
        \toprule
        \multirow{2}[1]{*}{Method} & \multicolumn{3}{c}{Ped. AP$_{IoU\geq0.5}$} & \multicolumn{3}{c}{Cycl. AP$_{IoU\geq0.5}$}
        \\ \cmidrule(lr){2-4} \cmidrule(lr){5-7}
        {} & Easy & Mod. & Hard & Easy & Mod. & Hard \\
        \midrule
        M3D-RPN~\cite{m3drpn} & 4.92 & 3.48 & 2.94 & 0.94 & 0.65 & 0.47 \\
        MonoPSR~\cite{monopsr} & 6.12 & 4.00 & 3.30 & \textcolor{red}{8.37} & \textcolor{red}{4.74} & \textcolor{red}{3.68} \\
        MonoPair~\cite{monopair} & 10.02 & 6.68 & 5.53 & 3.79 & 2.12 & 1.83 \\ \midrule
        Ours & \textcolor{red}{10.88} & \textcolor{red}{6.78} & \textcolor{red}{5.83} & 1.01 & 0.61 & 0.48 \\
        \bottomrule
    \end{tabular}}
    \end{center}
    \vspace{-1mm}
    \caption{\textbf{3D detection performance of Pedestrian and Cyclist categories} on KITTI official test set. \textcolor{red}{Red} indicates the best.}
    \label{tab:multiclass}
\end{table}

\section{Experiments}
\subsection{Dataset}
We evaluate the proposed model on the KITTI-Object benchmark~\cite{kitti}. It consists of 7481 training images and 7518 test images as well as the corresponding point clouds, comprising a total of 80256 labeled objects in eight classes. Each object is assigned to one of three difficulty levels according to truncation, occlusion and 2D box height. The training images are further split into 3712 training and 3769 validation images~\cite{3dproposal}. The official benchmark evaluates detection performance on three classes: \textit{Car}, \textit{Pedestrian} and \textit{Cyclist}. 
Evaluation metrics are based on precision-recall curves with 3D IoU threshold of 0.7 or 0.5. We adopt the official metric that computes the 40-point interpolated average precision (AP)~\cite{monodis}.

\subsection{Implementation Details}
\paragraph{2D Detector}
We use pre-trained Faster R-CNN~\cite{faster} with ResNet-101~\cite{resnet} as backbone. We adopt a six-level feature pyramid network~\cite{FPN}, in which an additional upsampled level is added.

\paragraph{Reconstruction Module} We set the dropout rate to 0.5 for 1D dropout layers and 0.2 for channel dropout layers. Network outputs (dimensions, NOCs) are normalized \wrt the mean and standard deviation calculated from training data. When training with multiple classes, we predict a set of class-specific latent vectors, dimensions and NOCs.

\paragraph{Data Augmentation}
During training, we apply random flip and photometric distortion augmentation. 
We set two NOC decoder branches at the last 1\texttimes1 convolutional layer for original and mirrored objects respectively.

\paragraph{Training Schedule}
The network is trained by the AdamW~\cite{adamw} optimizer with a weight decay of 0.01.
We take a batch size of 6 on two Nvidia RTX 2080 Ti GPUs,
and train the network using cosine learning rate decay with a base learning rate of 0.0002. Total epochs are 32 for the full training set and 50 for the split training set. For end-to-end training, we append a second cycle of 15 epochs with a reduced base learning rate of 0.00003.

\paragraph{Testing Configuration}
For epistemic uncertainty, we use 50 Monte Carlo dropout samples as in~\cite{kendall2017uncertainties}. By default, we only sample the global extractor to estimate the dimension uncertainty (see discussions in Section~\ref{ablation}). During post-processing, we use 3D non-maximum suppression (NMS) with an IoU threshold of 0.01.

\subsection{Comparison to the State of the Art}
We evaluate two variations of our model (fully self-supervised and plus LiDAR supervision) on the official test set and the validation split. Table \ref{tab:compare} lists the top methods from the official leaderboard. We observe that: (1) When trained \textbf{with LiDAR supervision}, our method outperforms the state of the art by a wide margin. Note that the top three competitors also use extra supervision and even extra data (by using depth estimators pre-trained on the much larger KITTI-Depth dataset). (2) When trained \textbf{without extra supervision}, our method still outperforms the non-depth methods (RTM3D~\cite{RTM3D}, MonoPair~\cite{monopair}) on the official test set. (3) Our method achieves state-of-the-art accuracy within a reasonable runtime (0.070 s, including Monte Carlo and PnP), whereas the top three competitors spend more than 0.2 s (not counting the 0.5 s depth estimation time of DORN~\cite{DORN}).

For Pedestrian and Cyclist, we present the detection performance in Table \ref{tab:multiclass}. Our method (with LiDAR supervision) achieves the best performance in Pedestrian detection, yet underperforms in Cyclist detection, presumably due to inadequate training samples.

\subsection{Ablation Studies} \label{ablation}
In this section, all models are trained and evaluated with the train/val split. We show the mean value of all six AP metrics (mAP) for 3D Car detection. All results are presented in Table~\ref{tab:ablation}.

\begin{figure*}[t]
  \begin{subfigure}{0.58\textwidth}
    \begin{center}
    \includesvg[width=0.95\linewidth]{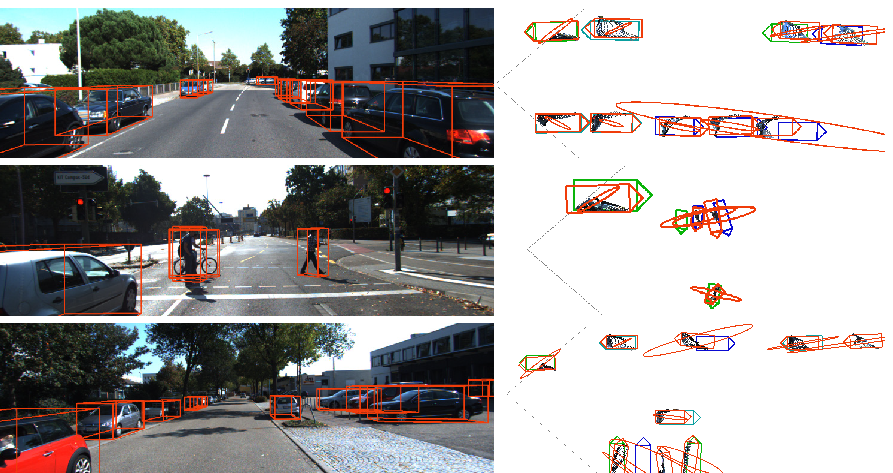}
    \end{center}
    \vspace*{-4.3mm}
    \caption{3D bounding boxes in images and bird's-eye views (BEV). \textcolor{box_red}{Red} indicates detected boxes (along with 95\% confidence ellipse), ground truth boxes are color coded by their occlusion levels: \textcolor{box_green}{fully visible}, \textcolor{box_cyan}{partly occluded}, \textcolor{box_blue}{difficult to see}.} \label{fig:viz_a}
  \end{subfigure}%
  \hspace*{\fill}
  \begin{subfigure}{0.39\textwidth}
    \begin{center}
    \includegraphics[width=0.80\linewidth]{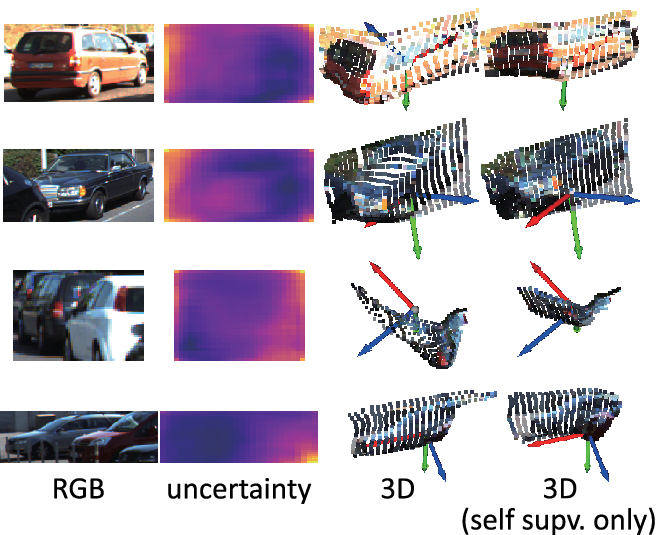}
    \end{center}
    \vspace*{-5.3mm}
    \caption{The image patches and their corresponding reconstruction results. Note that LiDAR helps regularizing the shape, which alleviates overfitting. Future work may explore other shape regularization techniques, \eg, using prior models.} \label{fig:viz_b}
  \end{subfigure}%
  \vspace*{1mm}
  \caption{\textbf{Visualization of detection and reconstruction results} on the KITTI validation set.} 
\label{fig:viz}
\end{figure*}

\paragraph{Self-Supervision versus LiDAR Supervision}
While self-supervision alone can achieve the state-of-the-art performance (28.57), using only LiDAR supervision leads to very poor performance (18.84), which demonstrates the importance of self-supervision. Nonetheless, the overall best result is reached with both supervisions together (31.21). As revealed in Fig.~\ref{fig:viz_b}, the self-supervised geometry does not necessarily match the true surface and is therefore prone to overfitting, which can be alleviated by the shape-regularizing effect from LiDAR supervision.

\begin{table}[t]
    \vspace*{-2mm}
    \begin{center}
    \scalebox{0.8}{%
    \setlength{\tabcolsep}{0.0em}
    \begin{tabular}{*{6}{>{\centering\arraybackslash}p{1.4cm}}}
        \toprule
        \makecell{$L_\text{proj}$ \\ (Self)} & \makecell{$L_\text{NOC}$ \\ (LiDAR)} & E2E & Epistemic & \makecell{Latent \\ vector} & mAP \\
        \midrule
        $L_\text{RKL}$ & \checkmark & {} & {} & \checkmark & 31.21 \\
        $L_\text{RKL}$ & {} & {} & {} & \checkmark & 28.57 \\
        {} & \checkmark & {} & {} & \checkmark & 18.84 \\ \midrule
        $L_\text{SmoothL1}$ & \checkmark & {} & {} & \checkmark & 26.35 \\
        $L_\text{LapKL}$ & \checkmark & {} & {} & \checkmark & 29.47 \\
        $L_\text{mKL}$ & \checkmark & {} & {} & \checkmark & 30.05 \\ \midrule
        $L_\text{LapKL}$ & \checkmark & \checkmark & {} & \checkmark & 29.73 \\ 
        $L_\text{RKL}$ & \checkmark & \checkmark & {} & \checkmark & 31.09 \\ \midrule
        $L_\text{RKL}$ & \checkmark & {} & dim & \checkmark & 31.47 \\
        $L_\text{RKL}$ & \checkmark & {} & full & \checkmark & 31.16 \\ \midrule
        $L_\text{RKL}$ & \checkmark & {} & {} & {} & 29.78 \\   
        \bottomrule
    \end{tabular}}
    \end{center}
    \vspace{-1mm}
    \caption{Results of \textbf{ablation studies} on reprojection loss function, LiDAR supervision, end-to-end training, epistemic uncertainty and latent vector.}
    \label{tab:ablation}
\end{table}

\paragraph{Robust KL Loss} By comparing the performance of smooth L1 loss (26.35), Laplacian KL loss (29.47), mixed KL loss (30.05) and Robust KL loss (31.21), we observe incremental improvements. The largest performance gap is between $L_\text{SmoothL1}$ and $L_\text{LapKL}$, which reveals the contribution of aleatoric uncertainty. Moreover, both Gaussian-Laplacian mixture and weight normalization make important contributions to the Robust KL loss, totaling an mAP increase of 1.74 compared to the Laplacian KL loss.

\paragraph{End-to-End Refinement} We observe that the performance of end-to-end refinement is strongly related to the baseline performance. For the baseline trained with Laplacian KL loss (29.47), end-to-end refinement boosts the mAP by 0.26 (29.73). For the baseline trained with Robust KL loss (31.21), however, end-to-end refinement slightly worsens the performance (31.09). This validates that self-supervised training with Robust KL loss can better optimize the network than end-to-end training via differentiable-PnP.

\paragraph{Epistemic Uncertainty} Estimating the epistemic uncertainty of box dimensions alone shows improvement to the baseline (31.47 vs 31.21). When sampling the full reconstruction network, we observe adverse effect on the detection performance (31.16).

\paragraph{Latent vector} When disabling the latent vector, the performance drops significantly (29.78 vs 31.21). To find out how the latent vector impacts the network performance, we evaluate the network sensitivity to the latent vector by resetting it to zero during inferring. As shown in Table~\ref{tab:latent}, only the network with aleatoric uncertainty is sensitive to the latent vector. This implies that the latent vector encodes very important information for the estimation of aleatoric uncertainty.

\begin{table}[t]
    \vspace*{-2mm}
    \begin{center}
    \scalebox{0.8}{%
    \setlength{\tabcolsep}{0.5em}
    \begin{tabular}{lccc}
        \toprule
        $L_\text{proj}$ & \makecell{mAP \\ (original)} & \makecell{mAP \\ (zero latent)} & Diff \\
        \midrule
        $L_\text{SmoothL1}$ (w/o aleatoric) & 26.35 & 26.33 & \textminus0.02\\
        $L_\text{RKL}$ (w/ aleatoric) & 31.21 & 29.23 & \textminus1.98\\
        \bottomrule
    \end{tabular}}
    \end{center}
    \vspace{-1mm}
    \caption{\textbf{Performance sensitivity to the latent vector}.}
    \label{tab:latent}
\end{table}

\subsection{Reliability of the Localization Uncertainty}
\begin{figure}
  \vspace*{-6.9mm}
  \begin{subfigure}[b]{0.5\linewidth}
    \includegraphics[width=\linewidth]{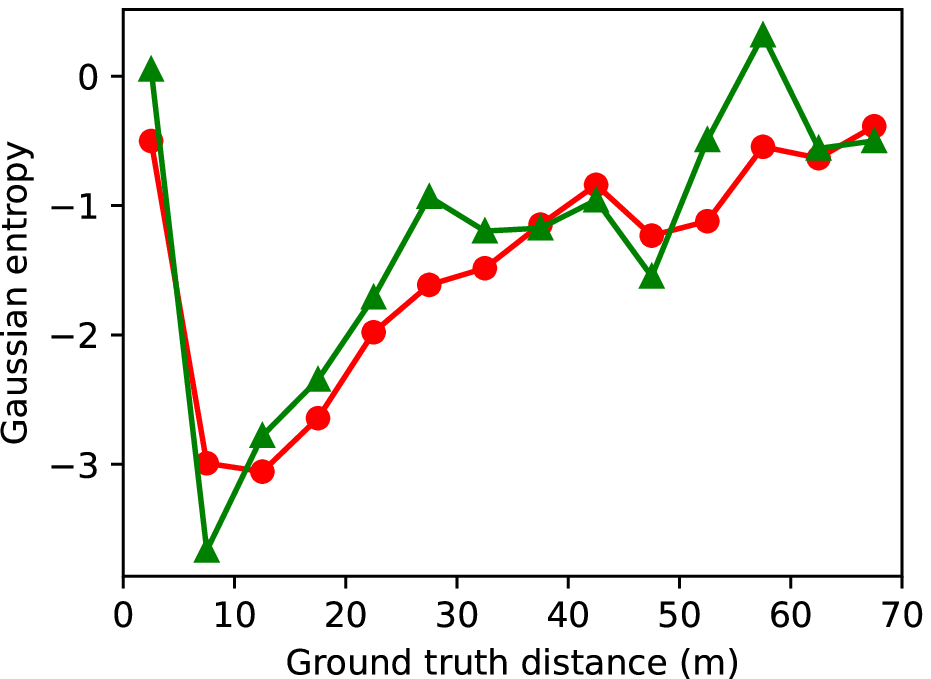}
    \vspace*{-5.5mm}
    \caption{Train split.} \label{fig:entropy_a}
  \end{subfigure}%
  \hspace*{\fill}
  \begin{subfigure}[b]{0.5\linewidth}
    \includegraphics[width=\linewidth]{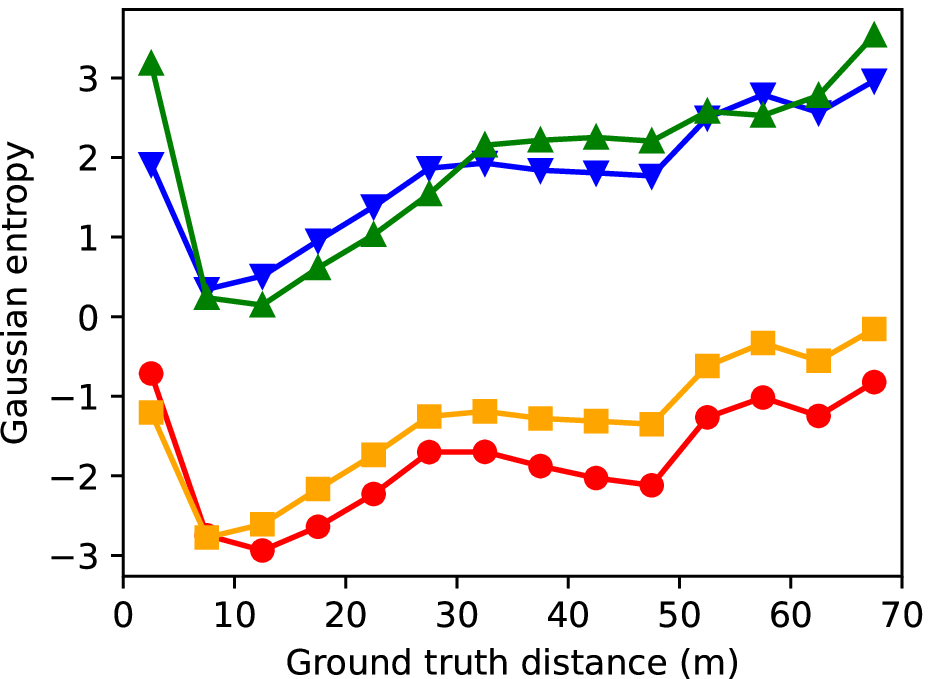}
    \vspace*{-5.5mm}
    \caption{Val split.} \label{fig:entropy_b}
  \end{subfigure}%
\caption{\textbf{Gaussian entropy of predicted and actual covariance matrices} vs object distance, tested on the Car category of train and val splits.} \label{fig:entropy}
\end{figure}

As the online covariance calibration is conducted on the training data, the model tends to be overconfident on the testing data. To assess the reliability of the localization uncertainty (\wrt the translation vector $\mathbf{t}$), we discretize the ground truth object distance $t_\text{z}$ into a number of bins. For the samples in each bin, we calculate the mean covariance matrix of prediction ($\overline{\mathbf{\Sigma}}_{\mathbf{t}^*}$), as well as the covariance matrix of the actual localization error ($\cov[\mathbf{t}^* - \mathbf{t}_\text{gt}]$). Ideally, these two covariance matrices should be equal. For comparison, we calculate the Gaussian entropy of the two covariance matrices respectively:
\begin{equation}
    H_\text{pred} = \frac{1}{2} \log \det \left( 2 \pi \text{e} \overline{\mathbf{\Sigma}}_{\mathbf{t}^*}\right),
\end{equation}
\begin{equation}
    H_\text{actual} = \frac{1}{2} \log \det \left( 2 \pi \text{e} \cov[\mathbf{t}^* - \mathbf{t}_\text{gt}] \right).
\end{equation}
As shown in Fig.~\ref{fig:entropy}, on the train split, the predicted uncertainty is very close to the actual error. This demonstrates the effectiveness of covariance calibration. For the unseen validation data, since epistemic sampling does not cover the full network, the model generally predicts overconfident results, which can be roughly corrected by applying an empirical covariance scaling factor.

\begin{figure}[t]
  \vspace*{-3mm}
\begin{center}
    \includegraphics[width=\linewidth]{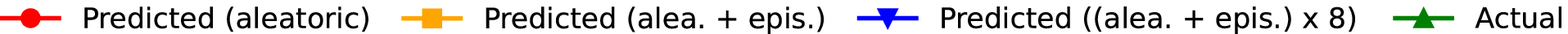}
\end{center}
\end{figure}

\section{Conclusion}
We presented the MonoRUn framework, a novel monocular 3D object detector with state-of-the-art performance and high practicality. To employ dense correspondence method for 3D detection in real driving scenes, we overcame the deficiency of geometry supervision by self-supervised reconstruction with uncertainty awareness. Meanwhile, we made uncertainty-aware deep regression networks easier to optimize by proposing the Robust KL loss. Finally, we are among the first to explore probabilistic 3D object localization by uncertainty propagation through PnP, which may open up new possibilities for downstream tasks such as robust tracking and motion prediction.

\paragraph{Acknowledgements}
This research was supported by the National Natural Science Foundation of China (Grant No.~52002285),
the Shanghai Pujiang Program (Grant No.~2020PJD075) and the Shenzhen Future Intelligent Network Transportation
System Industry Innovation Center (Grant No.~17092530321).


{\small
\bibliographystyle{ieee_fullname}
\bibliography{egbib}
}

\section{Supplementary Material}
As mentioned in the main paper, this supplementary material discusses the epistemic uncertainty, end-to-end training and Monte Carlo scoring in detail. Apart from that, it also provides the complete evaluation results on the official KITTI benchmark, including precision-recall plots.

\subsection{Details on Epistemic Uncertainty}
For the estimation of epistemic uncertainty, we adopt the Monte Carlo dropout approach described in \cite{kendall2017uncertainties}. By sampling the reconstruction network, we can estimate the epistemic uncertainty of object coordinates $\mathbf{x}^\text{OC}$. The predictive mean and variance are approximated by:
\begin{equation}
    \text{E}[\mathbf{x}^\text{OC}] \approx \overline{\mathbf{x}^\text{OC}} = \frac{1}{N_\text{MC}} \sum_i^{N_\text{MC}} \mathbf{x}^\text{OC}_i,
\end{equation}
\begin{equation}
    \var[\mathbf{x}^\text{OC}] \approx \frac{1}{N_\text{MC} - 1} \sum_i^{N_\text{MC}} ({\mathbf{x}^\text{OC}_i} - \overline{\mathbf{x}^\text{OC}})^2,
\end{equation}
where $\mathbf{x}^\text{OC}_i$ is the output of the $i$-th sampled network, $N_\text{MC}$ is the number of Monte Carlo samples (set to 50). 

Next, we need to transform the variances of 3D object coordinates into the variances of 2D reprojected coordinates. Strict 3D-2D variance projection requires knowing the object pose in advance, which is unavailable before the PnP module. Thus, we use the following approximation:
\begin{subequations}
\begin{empheq}[left=\empheqlbrace]{align}
    \var[u_\text{rp}^\text{norm}] & \approx \frac{1}{2} (\var[x^\text{OC}] + \var[z^\text{OC}]), \label{eqn:varu} \\
    \var[v_\text{rp}^\text{norm}] & \approx \var[y^\text{OC}],
\end{empheq}
\end{subequations}
where $u_\text{rp}^\text{norm}, v_\text{rp}^\text{norm}$ are the normalized reprojected coordinates (invariant to depth). Note that this approximation does not take object orientation into consideration, thus Eq.~\ref{eqn:varu} simply averages the horizontal variances.

Finally, the reconstruction module outputs the combined uncertainty of reprojected 2D coordinates:
\begin{subequations}
\begin{empheq}[left=\empheqlbrace]{align}
    \sigma_{u_\text{rp}^\text{norm}, \text{comb}}^2 & \approx \frac{1}{N_\text{MC}} \sum_i^{N_\text{MC}}{\sigma_{u_\text{rp}^\text{norm}, i}^2} + \var[u_\text{rp}^\text{norm}], \\
    \sigma_{v_\text{rp}^\text{norm}, \text{comb}}^2 & \approx \frac{1}{N_\text{MC}} \sum_i^{N_\text{MC}}{\sigma_{v_\text{rp}^\text{norm}, i}^2} + \var[v_\text{rp}^\text{norm}],
\end{empheq}
\end{subequations}
where $\sigma_{u_\text{rp}^\text{norm}, i}^2, \sigma_{v_\text{rp}^\text{norm}, i}^2$ is aleatoric uncertainty predicted by the $i$-th sampled network. It is to be observed that all the 2D variances above are in the normalized scale. The actual variances should be further multiplied by a scale factor $(f / t_\text{z})^2$, where $f$ is camera focal length in pixels and $t_\text{z}$ is the z-component of object pose. Since $t_\text{z}$ is unknown beforehand, we only apply this factor to the final pose covariance as a correction.

\subsection{Details on End-to-End Training}
End-to-end training is only investigated in the ablation studies, and is not included in our final training setup. Nevertheless, here we elaborate on the details of differentiable PnP and end-to-end training loss.

Regarding differentiable PnP, we generally follow the approach in BPnP~\cite{BPnP}, with the code completely re-implemented for higher efficiency and uncertainty awareness. The details are as follows.
\paragraph{Differentiating the PnP Algorithm}
The derivative of the PnP result $\mathbf{p}^*$ is as follows:
\begin{equation}
    \frac{\partial \mathbf{p}^*}{\partial(\cdot)^\text{T}}
    = -\mathbf{H}^{-1} \frac{\partial \mathbf{J}^\text{T} \mathbf{r}_\text{all}}
    {\partial(\cdot)^\text{T}} \Bigm|_{\mathbf{p}^*},
\end{equation}
where $(\cdot)$ stands for PnP inputs $\mathbf{x}^\text{OC}$ and $\boldsymbol{\sigma}$. 

\begin{proof}
Recall the MLE of object pose:
\begin{align}
\mathbf{p}^* &= 
\argmin_\mathbf{p} \frac{1}{2} \negthickspace \negthickspace \negthickspace \smashoperator[r]{\sum_{(u, v) \in RoI}}
\mathbf{r}_{(u, v)}^\text{T} \mathbf{\Sigma}_{(u, v)}^{-1} \mathbf{r}_{(u, v)} \notag\\
&= \argmin_\mathbf{p} \frac{1}{2} \mathbf{r}_\text{all}^\text{T} \mathbf{r}_\text{all}. 
\label{eqn:mlefix}
\end{align}
The gradient of the NLL function ($\dfrac{1}{2} \mathbf{r}_\text{all}^\text{T} \mathbf{r}_\text{all}$) \wrt $\mathbf{p}$ is as follows:
\begin{equation}
    \mathbf{g} = \mathbf{J}^\text{T} \mathbf{r}_\text{all}
\end{equation}
with $\mathbf{J} = \dfrac{\partial \mathbf{r}_\text{all}}{\partial \mathbf{p}^\text{T}}$.
When the optimization (Eq.~\ref{eqn:mlefix}) converges to $\mathbf{p}^*$, the gradient always satisfies
\begin{equation}
    \mathbf{g}_{\mathbf{p}^*} = \mathbf{0}.
\end{equation}
Therefore, the total derivative of $\mathbf{g}$ \wrt any PnP input equals zero:
\begin{equation}
    \frac{\mathrm{D} \mathbf{g}_{\mathbf{p}^*}}{\mathrm{D} (\cdot)^\text{T}} = \frac{\partial \mathbf{g}_{\mathbf{p}^*}}{\partial {{\mathbf{p}}^*}^\text{T}} \frac{\partial \mathbf{p}^*}{\partial (\cdot)^\text{T}} + \frac{\partial \mathbf{g}_{\mathbf{p}^*}}{\partial (\cdot)^\text{T}} = \mathbf{0},
\end{equation}
which implies that
\begin{equation}
    \frac{\partial \mathbf{p}^*}{\partial (\cdot)^\text{T}} = -\left( \frac{\partial \mathbf{g}_{\mathbf{p}^*}}{\partial {{\mathbf{p}}^*}^\text{T}} \right)^{-1} \frac{\partial \mathbf{g}_{\mathbf{p}^*}}{\partial (\cdot)^\text{T}}
    = -\mathbf{H}^{-1} \frac{\partial \mathbf{J}^\text{T} \mathbf{r}_\text{all}}
    {\partial(\cdot)^\text{T}} \Bigm|_{\mathbf{p}^*} \negmedspace.
\end{equation}
\end{proof}

\paragraph{Implementation Details of PnP}
The PnP forward process is implemented with Ceres Solver on CPU, while the backward process is implemented with the Autograd package of PyTorch on GPU. With our efficient implementation, the backward overhead of computing the exact second derivatives is negligible for training.

\paragraph{End-to-End Training Loss}
Leveraging the differentiable PnP, we apply smooth L1 loss on the Euclidean errors of estimated translation vector $\mathbf{t}^*$ and yaw angle $\beta^*$:
\begin{equation}
    L_\text{trans} = L_\text{SmoothL1}\left(\left\| \mathbf{t}^* - \mathbf{t}_\text{gt} \right\| \right),
\end{equation}
\begin{equation}
    L_\text{rot} = L_\text{SmoothL1}\left(\left\|
    \begin{bmatrix} 
        \cos{\beta^*} \\ 
        \sin{\beta^*}
    \end{bmatrix} - 
        \begin{bmatrix} 
        \cos{\beta_\text{gt}} \\ 
        \sin{\beta_\text{gt}}
    \end{bmatrix}
    \right\| \right).
\end{equation}
The overall loss for end-to-end training is:
\begin{equation}
\begin{aligned}
    L_\text{e2e} = & L_\text{2D} + L_\text{dim} + L_\text{score} + \lambda L_\text{calib} \\
    & + L_\text{trans} + L_\text{rot} + L_\text{NOC},
\end{aligned}
\end{equation}
where the reprojection loss is replaced by translation and rotation losses, and the NOC loss is added as regularization.

\subsection{Details on Monte Carlo Scoring}
By sampling $\mathbf{p}_i$ from the distribution $\mathcal{N}(\mathbf{p}^*, \mathbf{\Sigma}_{\mathbf{p}^*})$, the 3D localization score can be computed using Monte Carlo integration:
\begin{equation}
    c_\text{3DL} = \frac{1}{N_\text{MC}} \sum_{i=1}^{N_\text{MC}}
    f\left(IoU_\text{3D}\left(
    \begin{bmatrix}
        \mathbf{p}^* \\ \mathbf{d}
    \end{bmatrix}, 
    \begin{bmatrix}
    \mathbf{p}_i \\ \mathbf{d}
    \end{bmatrix}\right) \negmedspace \right),
\end{equation}
where vectors in the form $\begin{bmatrix} \mathbf{p} & \mathbf{d} \end{bmatrix}^\text{T}$ represent 3D boxes used for computing 3D IoU; $f(\cdot)$ is technically a step function with a hard IoU threshold, which is 1 for $IoU_\text{3D} \geq IoU_\text{threshold}$ and 0 otherwise. In practice, we use the clamped linear function:
\begin{equation}
    f(IoU_\text{3D}) = \max(0, \min(1, 2 IoU_\text{3D} - 0.5)).
\end{equation}

Table~\ref{tab:scoring} illustrates the performance of Monte Carlo and MLP scoring methods, along with the baseline of using only 2D detection score. We observe that, although Monte Carlo scoring proves effective compared to the baseline, its performance is still much lower than the MLP scoring network. This validates that fusing both pose uncertainty and network feature can produce a more reliable confidence score.
\begin{table}[h]
    \begin{center}
    \scalebox{0.9}{%
    \setlength{\tabcolsep}{1.0em}
    \begin{tabular}{lc}
        \toprule
        Scoring Method & mAP \\
        \midrule
        2D score only & 25.40 \\
        Monte Carlo & 28.19 \\
        MLP & 31.47 \\
        \bottomrule
    \end{tabular}}
    \end{center}
    \caption{Comparison between different scoring methods, based on the evaluation results on KITTI validation set.}
    \label{tab:scoring}
\end{table}

\subsection{Complete Evaluation Results on the Official KITTI Benchmark}
\begin{figure}[h]
  \begin{subfigure}[b]{0.5\linewidth}
    \includegraphics[width=\linewidth]{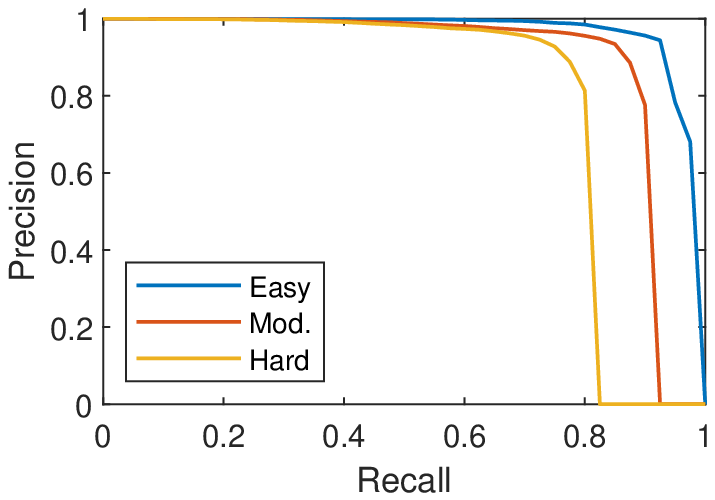}
    \vspace*{-6mm}
    \caption{2D detection.}
  \end{subfigure}%
  \hfill
  \begin{subfigure}[b]{0.5\linewidth}
    \includegraphics[width=\linewidth]{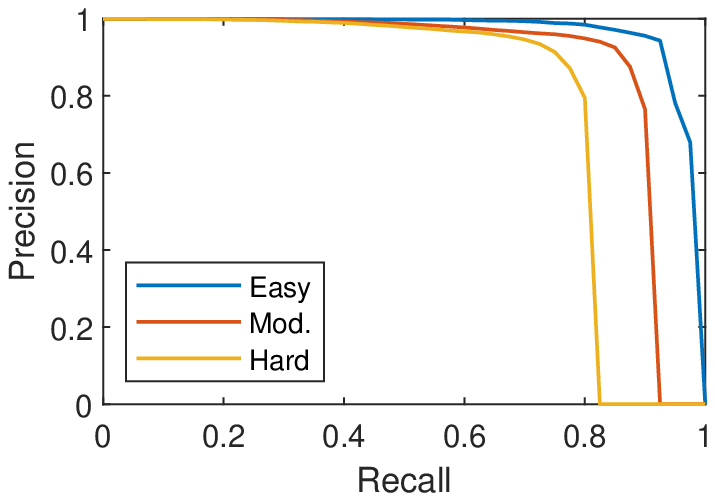}
    \vspace*{-6mm}
    \caption{Orientation similarity.}
  \end{subfigure}%
  \vskip\baselineskip
  \vspace*{-2.5ex}
  \begin{subfigure}[b]{0.5\linewidth}
    \includegraphics[width=\linewidth]{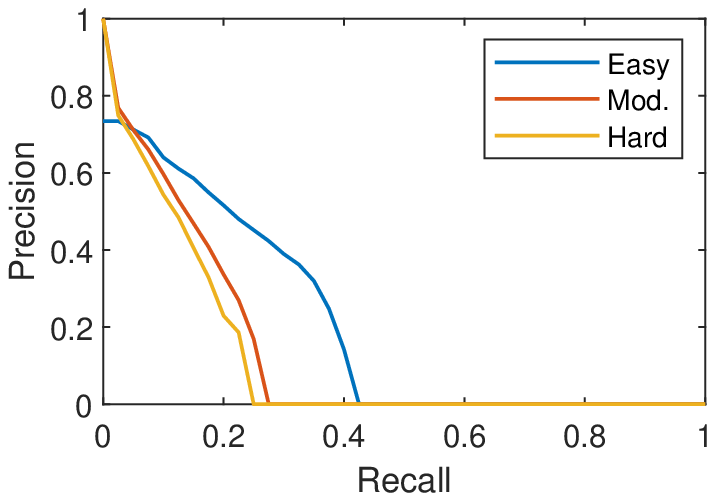}
    \vspace*{-6mm}
    \caption{3D detection.}
  \end{subfigure}%
  \hfill
  \begin{subfigure}[b]{0.5\linewidth}
    \includegraphics[width=\linewidth]{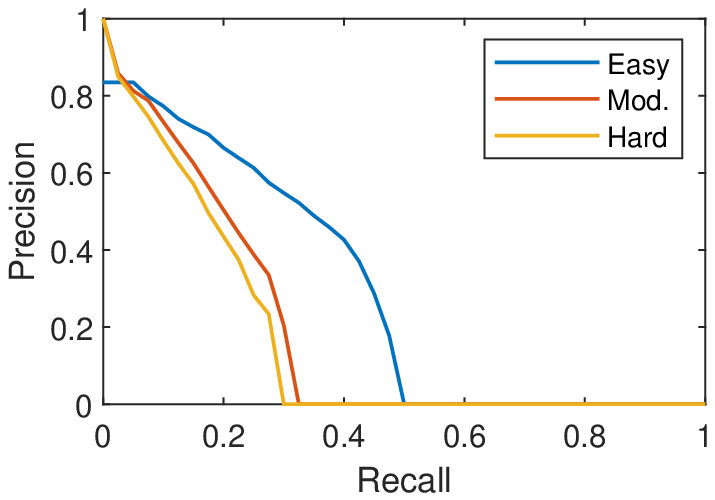}
    \vspace*{-6mm}
    \caption{BEV detection.}
  \end{subfigure}%
  \vspace*{-0.5ex}
\caption{Car, with LiDAR supervision.} \label{fig:car}
\end{figure}

\begin{figure}[h]
  \begin{subfigure}[b]{0.5\linewidth}
    \includegraphics[width=\linewidth]{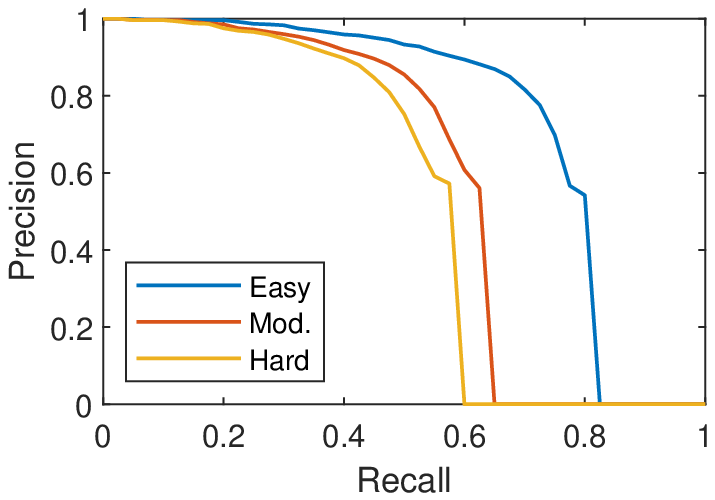}
    \vspace*{-6mm}
    \caption{2D detection.}
  \end{subfigure}%
  \hfill
  \begin{subfigure}[b]{0.5\linewidth}
    \includegraphics[width=\linewidth]{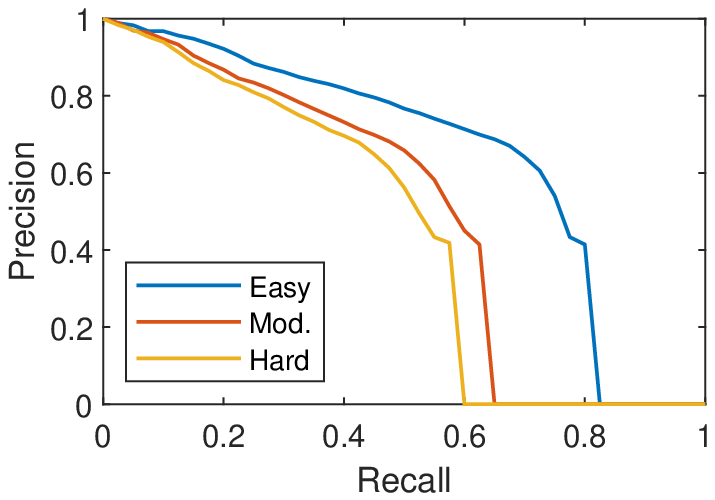}
    \vspace*{-6mm}
    \caption{Orientation similarity.}
  \end{subfigure}%
  \vskip\baselineskip
  \vspace*{-2.5ex}
  \begin{subfigure}[b]{0.5\linewidth}
    \includegraphics[width=\linewidth]{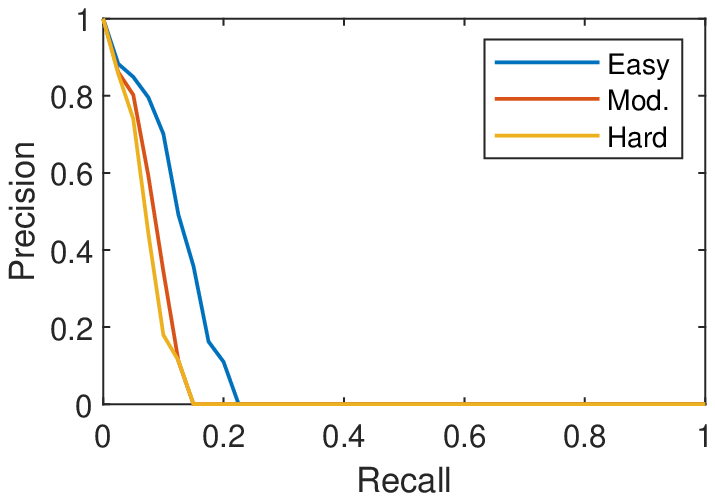}
    \vspace*{-6mm}
    \caption{3D detection.}
  \end{subfigure}%
  \hfill
  \begin{subfigure}[b]{0.5\linewidth}
    \includegraphics[width=\linewidth]{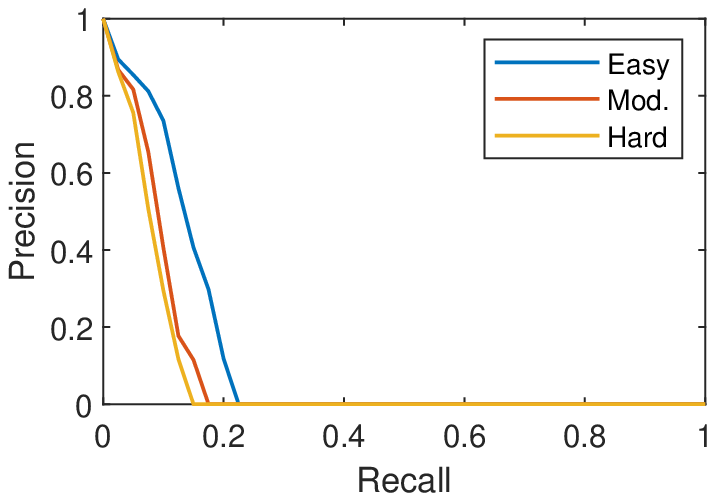}
    \vspace*{-6mm}
    \caption{BEV detection.}
  \end{subfigure}%
  \vspace*{-0.5ex}
\caption{Pedestrian, with LiDAR supervision.} \label{fig:ped}
\end{figure}

\begin{figure}[h]
  \begin{subfigure}[b]{0.5\linewidth}
    \includegraphics[width=\linewidth]{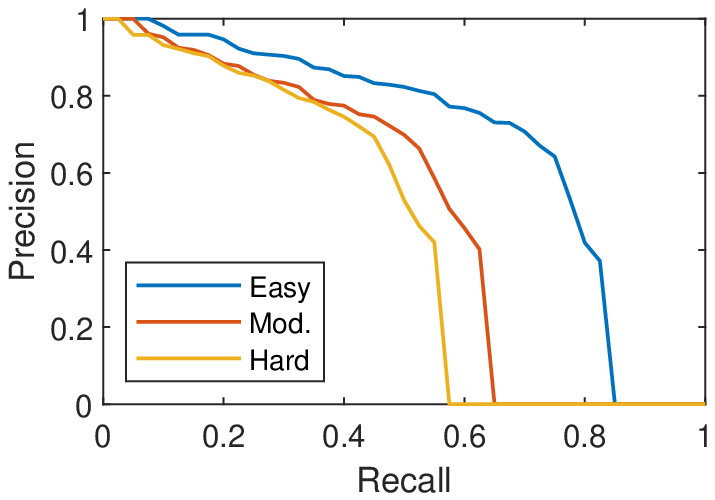}
    \vspace*{-6mm}
    \caption{2D detection.}
  \end{subfigure}%
  \hfill
  \begin{subfigure}[b]{0.5\linewidth}
    \includegraphics[width=\linewidth]{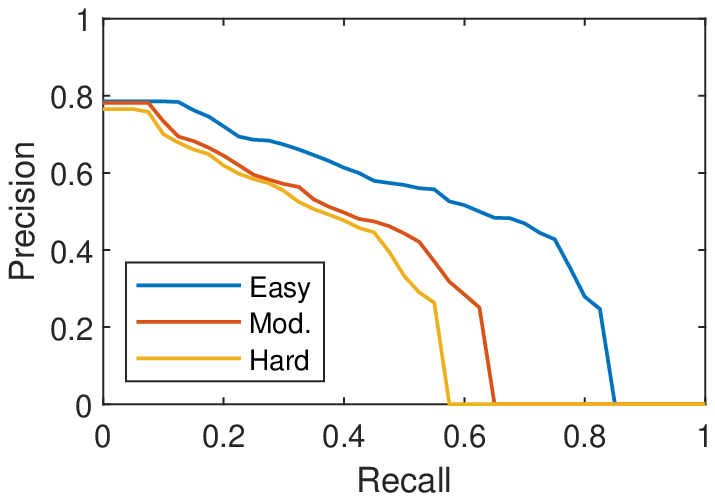}
    \vspace*{-6mm}
    \caption{Orientation similarity.}
  \end{subfigure}%
  \vskip\baselineskip
  \vspace*{-2.5ex}
  \begin{subfigure}[b]{0.5\linewidth}
    \includegraphics[width=\linewidth]{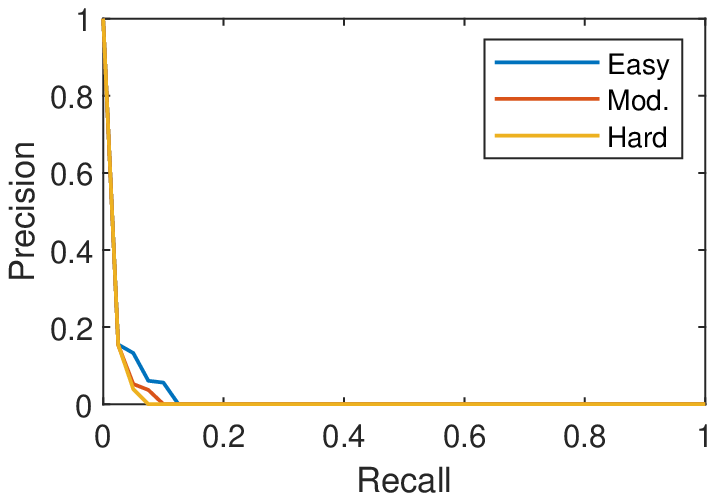}
    \vspace*{-6mm}
    \caption{3D detection.}
  \end{subfigure}%
  \hfill
  \begin{subfigure}[b]{0.5\linewidth}
    \includegraphics[width=\linewidth]{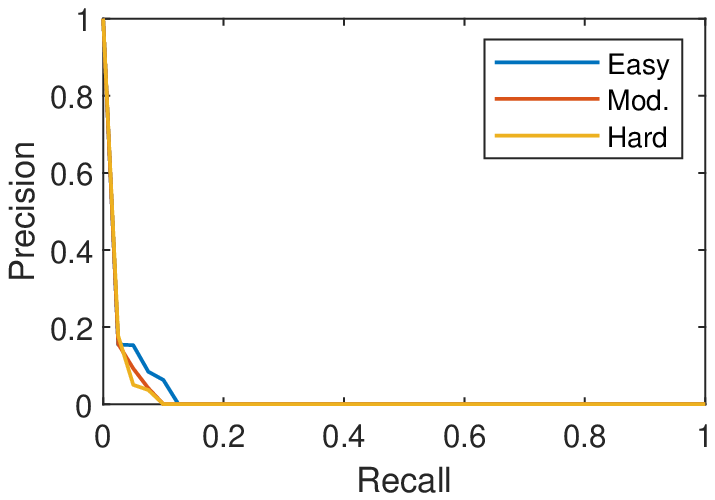}
    \vspace*{-6mm}
    \caption{BEV detection.}
  \end{subfigure}%
  \vspace*{-0.5ex}
\caption{Cyclist, with LiDAR supervision.} \label{fig:cyc}
\end{figure}

\begin{figure}[h]
  \begin{subfigure}[b]{0.5\linewidth}
    \includegraphics[width=\linewidth]{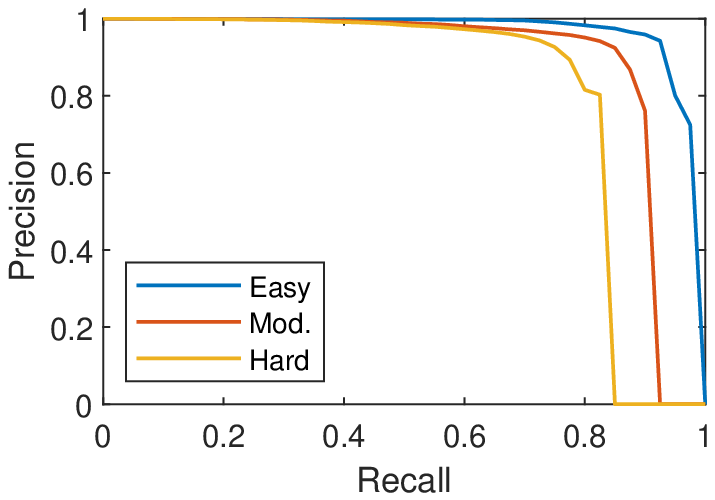}
    \vspace*{-6mm}
    \caption{2D detection.}
  \end{subfigure}%
  \hfill
  \begin{subfigure}[b]{0.5\linewidth}
    \includegraphics[width=\linewidth]{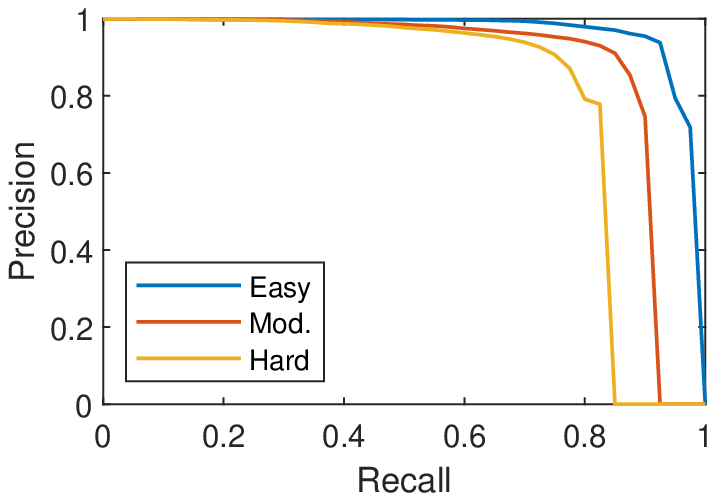}
    \vspace*{-6mm}
    \caption{Orientation similarity.}
  \end{subfigure}%
  \vskip\baselineskip
  \vspace*{-2.5ex}
  \begin{subfigure}[b]{0.5\linewidth}
    \includegraphics[width=\linewidth]{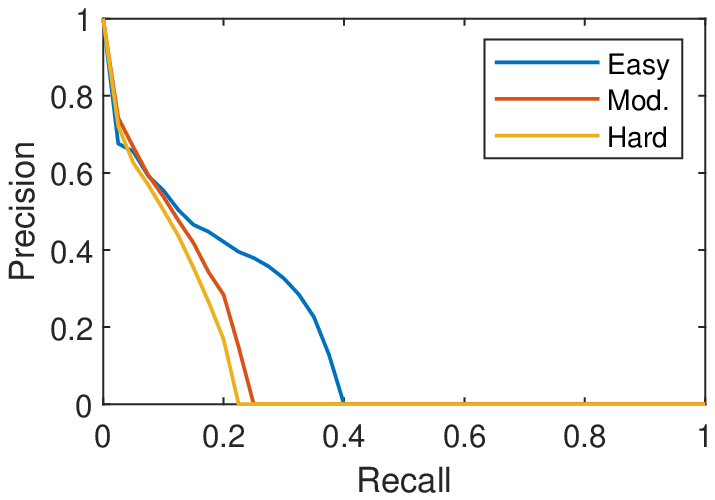}
    \vspace*{-6mm}
    \caption{3D detection.}
  \end{subfigure}%
  \hfill
  \begin{subfigure}[b]{0.5\linewidth}
    \includegraphics[width=\linewidth]{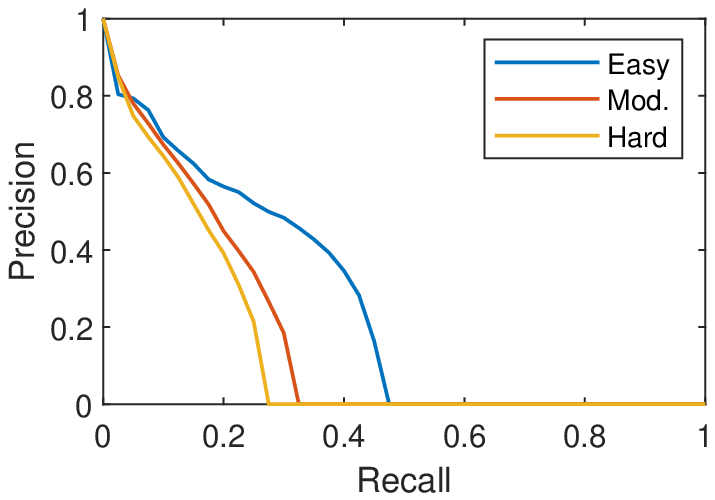}
    \vspace*{-6mm}
    \caption{BEV detection.}
  \end{subfigure}%
  \vspace*{-0.5ex}
\caption{Car, without LiDAR supervision (Fully self-supervised reconstruction).} \label{fig:car_self}
\end{figure}

\begin{figure}[t]
  \begin{subfigure}[b]{0.5\linewidth}
    \includegraphics[width=\linewidth]{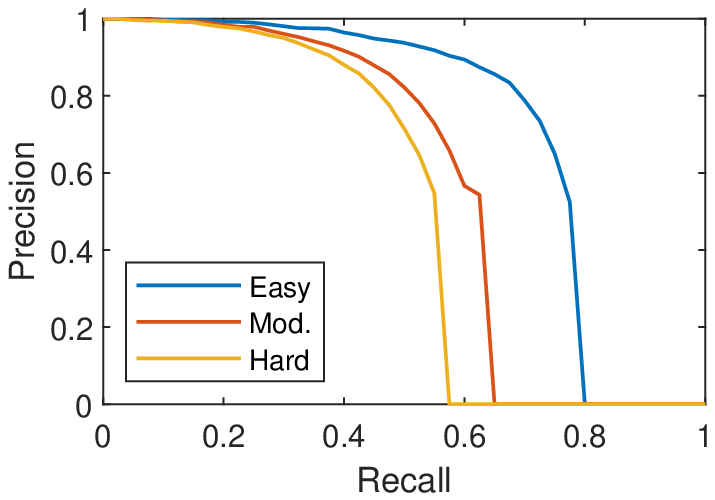}
    \vspace*{-6mm}
    \caption{2D detection.}
  \end{subfigure}%
  \hfill
  \begin{subfigure}[b]{0.5\linewidth}
    \includegraphics[width=\linewidth]{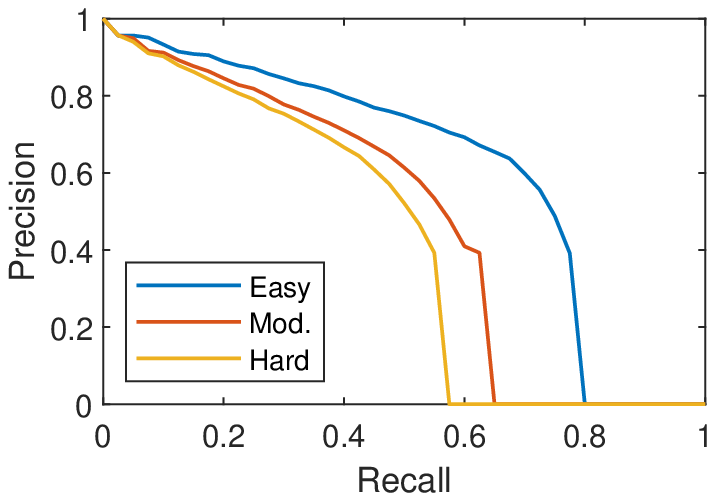}
    \vspace*{-6mm}
    \caption{Orientation similarity.}
  \end{subfigure}%
  \vskip\baselineskip
  \vspace*{-2.5ex}
  \begin{subfigure}[b]{0.5\linewidth}
    \includegraphics[width=\linewidth]{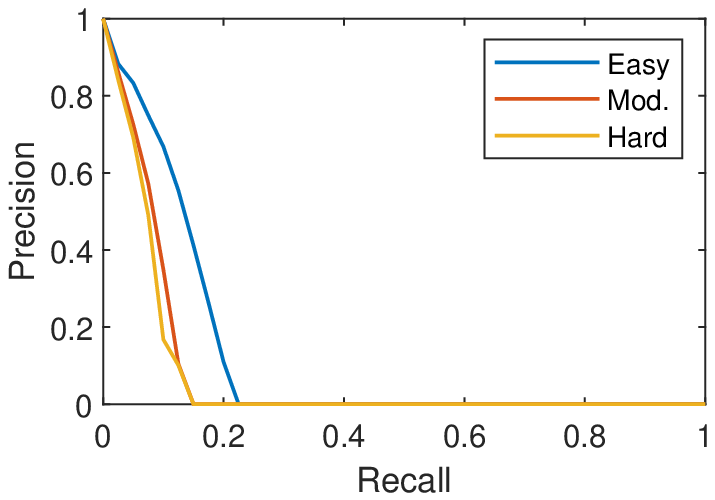}
    \vspace*{-6mm}
    \caption{3D detection.}
  \end{subfigure}%
  \hfill
  \begin{subfigure}[b]{0.5\linewidth}
    \includegraphics[width=\linewidth]{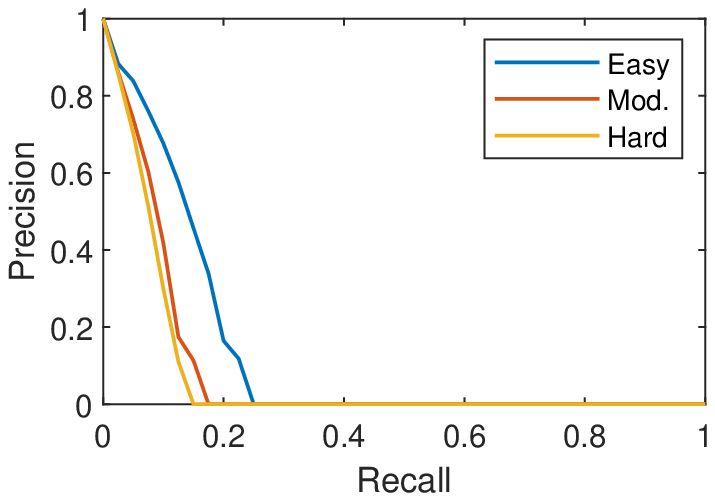}
    \vspace*{-6mm}
    \caption{BEV detection.}
  \end{subfigure}%
  \vspace*{-0.5ex}
\caption{Pedestrian, without LiDAR supervision (Fully self-supervised reconstruction).} \label{fig:ped_self}
\end{figure}

\begin{figure}[t]
  \begin{subfigure}[b]{0.5\linewidth}
    \includegraphics[width=\linewidth]{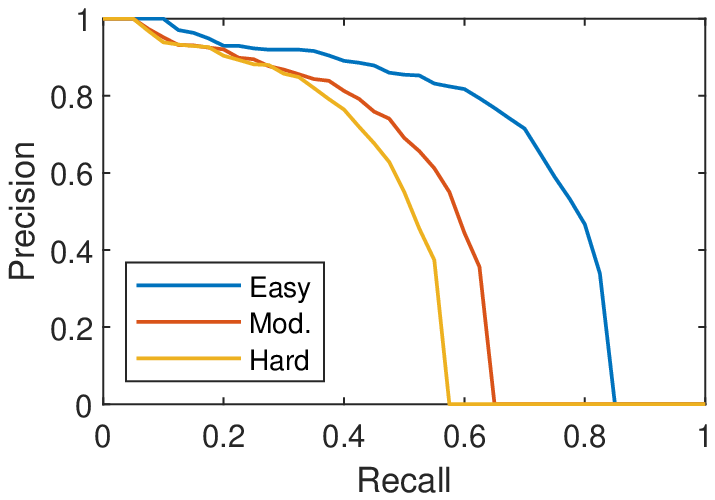}
    \vspace*{-6mm}
    \caption{2D detection.}
  \end{subfigure}%
  \hfill
  \begin{subfigure}[b]{0.5\linewidth}
    \includegraphics[width=\linewidth]{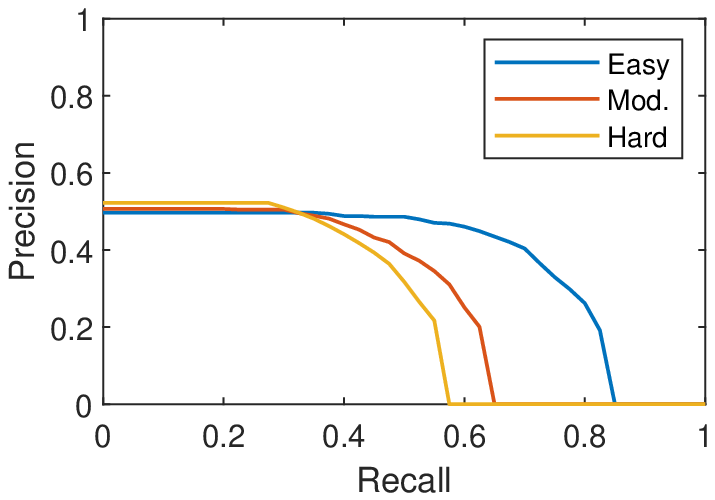}
    \vspace*{-6mm}
    \caption{Orientation similarity.}
  \end{subfigure}%
  \vskip\baselineskip
  \vspace*{-2.5ex}
  \begin{subfigure}[b]{0.5\linewidth}
    \includegraphics[width=\linewidth]{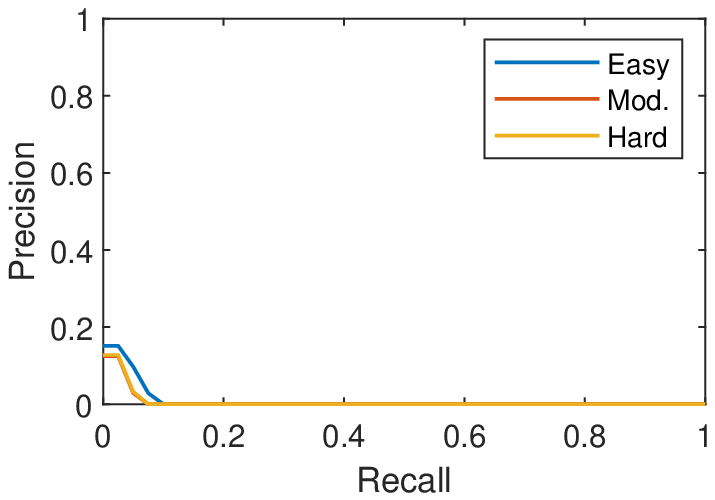}
    \vspace*{-6mm}
    \caption{3D detection.}
  \end{subfigure}%
  \hfill
  \begin{subfigure}[b]{0.5\linewidth}
    \includegraphics[width=\linewidth]{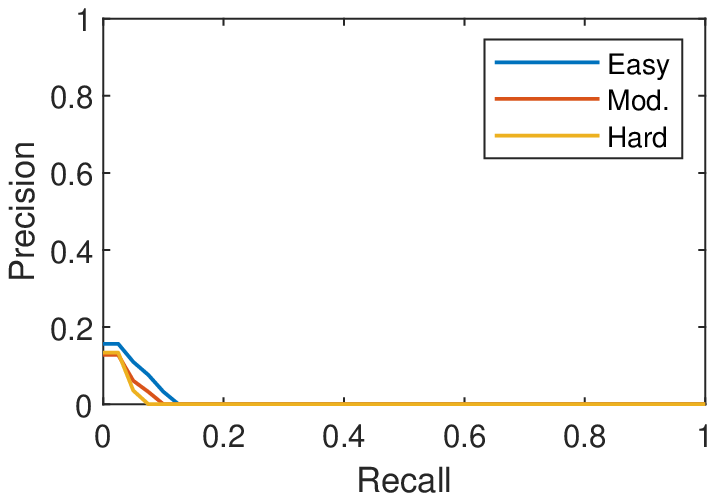}
    \vspace*{-6mm}
    \caption{BEV detection.}
  \end{subfigure}%
  \vspace*{-0.5ex}
\caption{Cyclist, without LiDAR supervision (Fully self-supervised reconstruction).} \label{fig:cyc_self}
\end{figure}

The official KITTI-Object benchmark consists of four tasks (2D detection, orientation similarity, 3D detection, BEV detection) on three classes (Car, Pedestrian, Cyclist). All metrics are based on the precision-recall curves given the IoU threshold or rotation threshold. We tested two variants of our model: with LiDAR supervision and without LiDAR supervision. The results are shown in \Cref{tab:lidar,tab:self} and \Cref{fig:car,fig:ped,fig:cyc,fig:car_self,fig:ped_self,fig:cyc_self}.

\vspace*{6ex}

\begin{table}[hb]
    \begin{center}
    \scalebox{0.9}{%
    \setlength{\tabcolsep}{1.0em}
    \begin{tabular}{lrrr}
        \toprule
        Benchmark & Easy & Mod. & Hard \\
        \midrule
        Car (2D det.) & 95.48 & 87.91 & 78.10 \\
        Car (orientation) & 95.44 & 87.64 & 77.75 \\
        Car (3D det.) & 19.65 & 12.30 & 10.58 \\
        Car (BEV det.) & 27.94 & 17.34 & 15.24 \\ \midrule
        Ped. (2D det.) & 73.05 & 56.40 & 51.40 \\
        Ped. (orientation) & 63.28 & 47.82 & 43.23 \\
        Ped. (3D det.) & 10.88 & 6.78 & 5.83 \\
        Ped. (BEV det.) & 11.70 & 7.59 & 6.34 \\ \midrule
        Cyclist (2D det.) & 67.47 & 49.13 & 43.41 \\
        Cyclist (orientation) & 49.04 & 34.36 & 30.22 \\
        Cyclist (3D det.) & 1.01 & 0.61 & 0.48 \\
        Cyclist (BEV det.) & 1.14 & 0.73 & 0.66 \\
        \bottomrule
    \end{tabular}}
    \end{center}
    \caption{With LiDAR supervision.}
    \label{tab:lidar}
\end{table}

\begin{table}[hb]
    \begin{center}
    \scalebox{0.9}{%
    \setlength{\tabcolsep}{1.0em}
    \begin{tabular}{lrrr}
        \toprule
        Benchmark & Easy & Mod. & Hard \\
        \midrule
        Car (2D det.) & 95.65 & 87.76 & 80.12 \\
        Car (orientation) & 95.48 & 87.33 & 79.51 \\
        Car (3D det.) & 16.04 & 10.53 & 9.11 \\
        Car (BEV det.) & 24.02 & 15.98 & 13.52 \\ \midrule
        Ped. (2D det.) & 71.27 & 55.80 & 49.47 \\
        Ped. (orientation) & 60.13 & 46.00 & 40.61 \\
        Ped. (3D det.) & 11.18 & 6.53 & 5.73 \\
        Ped. (BEV det.) & 12.03 & 7.27 & 6.20 \\ \midrule
        Cyclist (2D det.) & 68.85 & 50.32 & 44.36 \\
        Cyclist (orientation) & 37.29 & 27.95 & 25.27 \\
        Cyclist (3D det.) & 0.69 & 0.38 & 0.40 \\
        Cyclist (BEV det.) & 0.94 & 0.55 & 0.42 \\
        \bottomrule
    \end{tabular}}
    \end{center}
    \caption{Without LiDAR supervision (Fully self-supervised reconstruction).}
    \label{tab:self}
\end{table}

\end{document}